\documentclass[conference]{IEEEtran}
\IEEEoverridecommandlockouts
\usepackage{cite}
\usepackage{amsmath,amssymb,amsfonts}
\usepackage{algorithmic}
\usepackage{graphicx}
\usepackage{xcolor}
\usepackage{makecell}
\usepackage{booktabs} 
\usepackage{subfig}
\usepackage{amsmath}
\usepackage[normalem]{ulem}
\usepackage{enumitem}
\usepackage{multirow}
\usepackage[nomargin,inline,marginclue,draft]{fixme}
\usepackage{balance}
\usepackage{changepage}
\usepackage{bm}
\usepackage[colorlinks,linkcolor=blue]{hyperref}
\usepackage{setspace}
\usepackage{mathrsfs}
\usepackage{ulem}
\usepackage{verbatim}
\usepackage{diagbox}
\usepackage{url}

\newcommand{\rev}[1]{{#1}}

\newlength\savedwidth

\def\BibTeX{{\rm B\kern-.05em{\sc i\kern-.025em b}\kern-.08em
		T\kern-.1667em\lower.7ex\hbox{E}\kern-.125emX}}
		
\begin{document}
\title{DisenHCN: Disentangled Hypergraph Convolutional Networks for Spatiotemporal Activity Prediction}

\makeatletter
\newcommand{\linebreakand}{%
\end{@IEEEauthorhalign}
\hfill\mbox{}\par
\mbox{}\hfill\hspace{0.75cm}\begin{@IEEEauthorhalign}
}
\makeatother

\newcommand{\authordrift}{\hspace{0.85cm}}


	\author{\IEEEauthorblockN{Yinfeng~Li,
			Chen~Gao\textsuperscript{\textsection}, 
			Quanming~Yao,
			Tong~Li,
			Depeng~Jin,
			Yong~Li
		}
		\IEEEauthorblockA{Beijing National Research Center for Information Science and Technology\\
			Department of Electronic Engineering, Tsinghua University, Beijing 100084, China\\}
		liyf19@mails.tsinghua.edu.cn, chgao96@gmail.com, \{qyaoaa, tongli, jindp, liyong07\}@tsinghua.edu.cn
	} %


\maketitle
	
		\begingroup\renewcommand\thefootnote{\textsection}
	\footnotetext{Chen Gao is the Corresponding Author.}
	\endgroup
	
\begin{abstract}
Spatiotemporal activity prediction, aiming to predict user activities at a specific location and time, is crucial for applications like urban planning and mobile advertising. Existing solutions based on tensor decomposition or graph embedding suffer from the following two major limitations: 1) ignoring the fine-grained similarities of user preferences; 2) user's modeling is entangled. In this work, we propose a hypergraph neural network model called DisenHCN~(short for {\underline{Disen}tangled \underline{H}ypergraph \underline{C}onvolutional \underline{N}etworks}) to bridge the above gaps.
In particular, we first unify the fine-grained user similarity and the complex matching between user preferences and spatiotemporal activity into a heterogeneous hypergraph. We then 
disentangle the user representations into different aspects (location-aware, time-aware, and activity-aware) and aggregate corresponding aspect's features on the constructed hypergraph, capturing high-order relations from different aspects and disentangles the impact of each aspect for final prediction.
Extensive experiments show that our DisenHCN outperforms the state-of-the-art methods by 14.23\% to 18.10\% on four real-world datasets. Further studies also convincingly verify the rationality of each component in our DisenHCN.
\end{abstract}
	

\section{Introduction}
\label{sec::intro}

Over the past decades, the mobile Internet has dramatically changed the way people behave in daily life. People rely on their mobile Apps to perform daily activities and generate massive {\bf U}ser {\bf S}patio-{\bf T}emporal {\bf A}ctivitiess~(USTA) data, such as Twitter messages and App usage records, illustrated in Fig. 1(a). These USTA data are generally associated with user IDs, geographic information, and posting time, offering an excellent opportunity to accurately predict users' spatiotemporal activities~\cite{w4kdd13}. Under this circumstance, spatiotemporal activity prediction is widely recognized as a fundamental task for applications like urban
planning~\cite{WDGTC,crossmapwww17,zhang2016gmove,yuan2017pred,liu2020ACTOR} and mobile advertising~\cite{zheng2010UCLAF,MCTF_WWW2015,fan2019personalized,yu2020SAGCN,zhang2018crossmodal,li2020apps}.

Formally, the task of spatiotemporal activity prediction is defined to predict the specific activity of a user at a given time and location.
Early classical tensor-based methods~\cite{zheng2010UCLAF,MCTF_WWW2015,fan2019personalized,WDGTC} approached this task as a tensor-decomposition problem by representing USTA data as a four-dimensional tensor. 
However, these methods suffer from the limitation of small model capacity. To address it, graph-based models~\cite{crossmapwww17,yu2020SAGCN,liu2020ACTOR} were proposed, in which location, time, and activity are represented as three kinds of nodes while their relations are represented as heterogeneous edges.
\begin{figure}[t]
	\centering
	\includegraphics[width=0.48\textwidth]{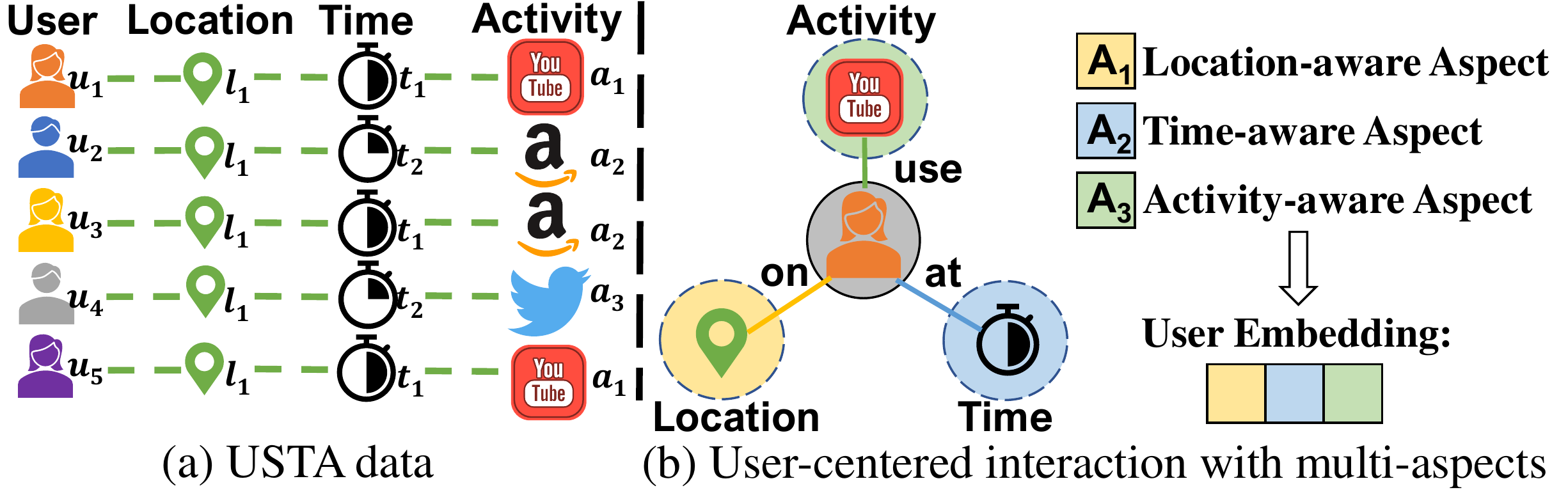}
	\setlength{\abovecaptionskip}{-0.2cm}
	\setlength{\belowcaptionskip}{-0.cm}
	\caption{The illustration of (a) the USTA data and (b) the different aspects of user preferences for the user representation.}
	\label{fig::user-centered}
\end{figure}

Although these graph-based methods have achieved more promising performance,
there are still unsolved critical challenges as follows.
\begin{itemize}[leftmargin=*]
	\item \textbf{Fine-grained user similarities are ignored.}
	The user similarity (similarity of user preference) is one of the most important information concealed in USTA data, since predicting a user's future behaviors requires collaborative learning of other users, similarly as collaborative filtering in recommender systems~\cite{he2017NCF,wang2020DGCF}.
	There are two major properties of user similarity that bring challenges to its modeling.
	First, user similarity has multiple types. 
	For example, two users visit the same location suggesting they may have similar location preferences, while two users have the same activity reflecting they may have similar tastes of activity.
	Second, different types have different granularities. For example, two users with the same activity at one same location reveal they may have similar preferences of both location and activity, which is a finer-grained similarity compared with the example we mentioned above.  
	However, existing works separately model time and location by roughly representing them as different types of nodes on the graph. 

\item \textbf{User's modeling is entangled.} 
In spatiotemporal activity prediction, user preferences focus on three aspects: location, time, and activity. 
In other words, user behaviors are driven by these three factors simultaneously. 
This property of multi-aspect preference distinguishes the task with 
traditional user-modeling problems, such as collaborate filtering (CF) in recommender systems~\cite{he2017NCF,he2020lightgcn}.
However, existing methods uniformly embed users into one same embedding space, 
which is insufficient to model the multi-aspect preferences that driven user behaviors.	
On the other hand, from the perspective of preference matching,
there exist a kind of {\em user-centered} one-to-many interaction. However, existing methods can only handle one-to-one interactions and fail to model the one-to-many interactions.
\end{itemize}

To address the above challenges, 
we propose a novel method, {\underline{\textbf{Disen}}tangled \underline{\textbf{H}}ypergraph \underline{\textbf{C}}onvolutional \underline{\textbf{N}}etworks} (DisenHCN).
Specifically, we first construct a heterogeneous hypergraph to encode the fine-grained user similarities and one-many complex matching between user and spatiotemporal activity.
We then project users to disentangled embeddings for decoupling preferences among location, time, and activity.
Efficient hypergraph convolution layers are then proposed on the hypergraph for embedding learning.
The independence of disentangled user embeddings is ensured by an independence-modeling module.

Our contributions can be summarized as follows.
\begin{itemize}[leftmargin=*]
\item To the best of our knowledge, 
we are the first to introduce 
heterogeneous hypergraph into the problem of spatiotemporal activity prediction.
We build multiple types of hyperedges, enabling the explicit modeling of the fine-grained user similarities. 

\item For the constructed heterogeneous hypergraph, we propose to disentangle user embeddings into several subspaces. We then propose efficient hypergraph convolution layers, 
including intra-type and inter-type aggregation on multi-type hyperedges, 
ensuring the disentangled user embeddings can learn the corresponding preferences on three aspects, \textit{i.e.},
location, time, and activity,
simultaneously.


\item We conduct extensive experiments on four real-world spatiotemporal activity datasets to demonstrate the superiority of our proposed model, significantly outperforming the existing state-of-the-art baselines by 14.23\% to 18.10\%. Further studies confirm the effectiveness of each designed component.
\end{itemize}

The remainder of this paper is as follows. We first review the related works in Section~\ref{sec::relatedwork} and provide some preliminaries in Section~\ref{sec::profdef}. We then introduce our proposed method in Section~\ref{sec::method} and conduct experiments in Section~\ref{sec::experiments}. Last we conclude this paper and discuss future works in Section~\ref{sec::conclusion}.

\section{Related Work}\label{sec::relatedwork}

\subsection{Spatiotemporal Activity Prediction}

Existing research works on spatiotemporal activity prediction~\cite{sizov2010geofolk,w4kdd13,yuan2017pred,zhang2016gmove} can be divided into two categories: the earlier {\em tensor-based} methods and recently-proposed {\em graph-based} ones.

\subsubsection{Tensor-based} 
These methods build a tensor with the USTA data and conduct tensor factorization~\cite{shashua2005tensorfact} to learn latent features. UCLAF~\cite{zheng2010UCLAF} \rev{builds} a user-location-activity tensor to learn latent representations for user, location, and activity. MCTF~\cite{MCTF_WWW2015} conducts tensor factorization on sparse user-generated data to perform multi-dimensional collaborative recommendations for Who (User), Where (Location), When (Time), and What (Activity). Fan {\em et al.}~\cite{fan2019personalized} combines tensor factorization and transfer learning to predict user's online activities. WDGTC~\cite{WDGTC} conducts tensor decomposition with graph laplacian penalty for passenger flow prediction.

\subsubsection{Graph-based}
To address tensor-based methods' limitation of model capacity,
graph-based methods are proposed in spatiotemporal activity prediction. CrossMap~\cite{crossmapwww17} and ACTOR~\cite{liu2020ACTOR} map different regions, hours, and activities into the same latent space by graph-embedding techniques.
CAP~\cite{chen2019cap}  predicts users' App usage with a heterogeneous graph embedding model. Recently, the emerging Graph Convolutional Network (GCN)~\cite{GCN} has injected new vitality into spatiotemporal activity prediction. 
SA-GCN~\cite{yu2020SAGCN} adopts GCN to obtain semantic-aware representations, achieving promising performance in App-usage prediction task.

\subsection{Hypergraph Learning}
\label{sec:rel:hyper}

Hypergraph~\cite{icml2006hypergraph} extends the normal graph by introducing a special edge, \textit{hyperedge}.
Hyperedge can connect two or more vertices ({\em i.e.}, nodes) to naturally model high-order relations in real-world scenarios. 
For example, in the social network, the high-order social relations among more than two users (vertices) can be modeled with hyperedges~\cite{yu2021MHCN}. 
Similar high-order relations including user-item~\cite{ji2020dual}, text-word~\cite{ding2020HyperGAT}, etc.
In short, compared with graph, hypergraph naturally possesses
the ability to model higher-order connections.
HGNN~\cite{feng2019HGNN} first extends graph convolution to hypergraph. HyperGAT~\cite{ding2020HyperGAT} extends graph attention to hypergraph for inductive text classification. HGC-RNN~\cite{yi2020HRCN} combines recurrent neural network and hypergraph to learn temporal dependency and high-order relations from the data sequence. Yuuki {\em et al.}~\cite{Pagerank} develop clustering algorithms based on PageRank for hypergraphs. 
Moreover, hypergraph has also been widely used in recommender systems~\cite{wang2020HyperRec,ji2020dual,LCFN,yu2021MHCN} to model the high-order correlations among users and items. 

Recently,
heterogeneous hypergraph~\cite{sun2021HWNN,zhu2016hete_doc,yang2019hete-social} further extends the capability of hypergraph~\cite{icml2006hypergraph} with multiple types of vertices and hyperedges in real-world applications such as document recommendation~\cite{zhu2016hete_doc} and location-based social relationship~\cite{yang2019hete-social}.

{\color{black}
In this work, we construct the heterogeneous hypergraph to capture the fine-grained user similarities and complex matching between user preferences and spatiotemporal activities.
}

\subsection{Disentangled Representation Learning}
Disentangled representations explore to separately modeling the multiple explanatory factors 
behind the data~\cite{bengio2013representation}. Some earlier works~\cite{burgess2018beta-VAE,chen2018VAE-ISO,higgins2016beta,kim2018disentangling} derive various regularization terms for variational auto-encoders~\cite{kingma2013VAE} from an information-theoretic perspective. Recently, several methods~\cite{ma2019disenrec-nips,wang2020DGCF,wang2020disenhan,zheng2021DICE} are proposed for disentangling user embeddings in graph-based models. 
In this work, different from them, we disentangle the user embeddings in the hypergraph to model user behaviors from three aspects of the preferences: location, time, and activity.

\section{Preliminary}
\label{sec::profdef}
\begin{table}[t]
	\centering
	\caption{\rev{The description of commonly used notations.}}
	\setlength\tabcolsep{4pt}
	\scalebox{0.95}{
	\begin{tabular}{c|l}
		\toprule[1pt]
		\textbf{\rev{Notations}} & \textbf{\rev{Description}}\\ \hline
		\rev{$\mathcal{D}$} & \rev{The set of spatiotemporal activity records.} \\ \hline
		\rev{$\mathcal{U}$ / $N_U$} & \rev{The set of users / the number of users.} \\ \hline
		\rev{$\mathcal{L}$ / $N_L$} & \rev{The set of locations / the number of locations.} \\ \hline
		\rev{$\mathcal{T}$ / $N_T$} & \rev{The set of time-slots / the number of time-slots.} \\ \hline
		\rev{$\mathcal{A}$ / $N_A$} & \rev{The set of activities / the number of activities.} \\ \hline
		\rev{$u$ / $\mathbf{p}_u$} & \rev{User ID / User embedding.}\\ \hline
		\rev{$l$ / $\mathbf{q}_l$} & \rev{Location ID / Location embedding.}\\ \hline
		\rev{$t$ / $\mathbf{r}_t$} & \rev{Time ID / Time embedding.}\\ \hline
		\rev{$a$ / $\mathbf{s}_a$} & \rev{Activity ID / Activity embedding.}\\ \hline
		\rev{$f(\cdot)$} & \rev{The model for spatiotemporal activity prediction.}\\ \hline
		\rev{$\hat{y}_{ulta}$} & \rev{The predicted probability with given $u,l,t,a$.}\\ \hline
		\rev{${\mathcal G}$} & \rev{Heterogeneous hypergraph.} \\ \hline
		\rev{$e_i$} & \rev{Hyperedge with ID $i$.} \\ \hline
		\rev{$\mathcal{V}$ / $\mathcal{T}_v$} & \rev{Node set / Node type.} \\ \hline
		\rev{$\mathcal{E}$ / $\mathcal{T}_e$} & \rev{Hyperedge set / Hyperedge type.} \\ \hline
		\rev{$\mathcal{T}_e^u$} & \rev{Hyperedge type for user similarities.} \\ \hline
		\rev{$\mathcal{S}$} & \rev{Set of disentangled subspaces.}\\ \hline
		\rev{$\mathbf{p}_{u,s}$} & \rev{Disentangled embedding chunk of user $u$.}\\ \hline
		\rev{$\sigma(\cdot)$} & \rev{The Sigmoid function.}\\ \hline
		\rev{$d$} & \rev{The embedding size.}\\ \hline
	    \rev{$\ell$} & \rev{The $\ell$-th layer.}\\ \hline
	    \rev{$L$} & \rev{The number of layers.}\\ \hline
		\rev{$\mathrm{AGG}$} & \rev{The aggregation function.}\\
		\hline
		\rev{$\mathbf{a}$} & \rev{The trainable attention vector.} \\ \hline
		\rev{$F$} & \rev{The symbol of Eff-HGConv.} \\ \hline
		\rev{$\mathbf{W}$} & \rev{The transformation matrix.} \\ \hline
		\rev{$\mathbf{b}$} & \rev{The transformation bias.} \\ \hline
		\rev{$\mathcal{L}_{IND}$} & \rev{The independence loss function.} \\ \hline
		\rev{$\text{dCov}(\cdot)$} & \rev{The distance covariance.} \\ \hline
		\rev{$\text{dVar}(\cdot)$} & \rev{The distance variance.} \\ \hline
		\rev{$\mathcal{L}_{BPR}$} & \rev{The BPR loss.}\\ \hline
		\rev{$\mathcal{O}$} & \rev{The pairwise training set.}\\ \hline
		\rev{$\lambda$} & \rev{The $L_2$ regularization coefficient.}\\ \hline
		\rev{$\bf\Theta$} & \rev{The model parameters.}\\ \hline
		\rev{$\mathcal{L}$} & \rev{The total loss.}\\ \hline
		\rev{$\gamma$} & \rev{The independent modeling coefficient.}\\
\bottomrule[1pt]
	\end{tabular}}
	\label{tab::symbal}
\end{table}
\subsection{Problem Formulation}
Let~$\mathcal{D}=\{(u_i, l_i,t_i,a_i) \}$ denote a corpus of the quadruple spatiotemporal activity records represented in Fig.~\ref{fig::user-centered} (a), where $u_i,l_i,t_i,a_i$ represent user ID, location ID, time-slot ID and activity ID, respectively\footnote{We use the same symbol $i$ to present the different indexes of four dimensions to avoid too many symbols.}. 
We use $\mathcal{U},\mathcal{L},\mathcal{T},\mathcal{A}$ to denote the sets of users, locations, time-slots and activities respectively, and use $N_U,N_L,N_T,N_A$ to represent the size of these sets. Hence, the task of spatiotemporal activity prediction can be formulated as follows:\\
\textbf{Input}:~The observed spatiotemporal activity records $\mathcal{D}$.\\
\textbf{Output:}~A model to estimate the probability that a user $u$ at location $l$ and time-slot $t$ will conduct the target activity $a$, formulated as~$\hat{y}_{ulta}=f(u,l,t,a)$.
\rev{Note that spatiotemporal activity is recorded as a quadruple data $(u, l, t, a)$ in our task and the model $f(\cdot)$ measures the probability that $(u, l, t, a)$ occurs. In other words, $(u, l, t, a)$ in training set is observed record and ``$a$'' represents the known input; while in the test set, ``$a$'' represents the activity to be predicted.}

\subsection{Heterogeneous Hypergraph}
\label{sec::hyper_def}

Formally, a heterogeneous hypergraph~\cite{sun2021HWNN,zhu2016hete_doc,yang2019hete-social}
can be formulated as ${\mathcal G}=({\mathcal V},{\mathcal E}, \mathcal{T}_v, \mathcal{T}_e)$ with a vertex-type mapping function $\phi:\mathcal{V}\longmapsto \mathcal{T}_v$ and a hyperedge-type mapping function $\psi:\mathcal{E}\longmapsto \mathcal{T}_e$. Note that each vertex $v\in \mathcal{V}$ and hyperedge $e\in \mathcal{E}$ belong to one particular type. 
Here $\mathcal{V}$ is the set of vertices, and $\mathcal{T}_v$ is the set of vertex-types; 
$\mathcal{E}$ is the set of hyperedges, and $\mathcal{T}_e$ is set of the hyperedge-types. Hypergraphs allow more than two nodes to be connected by one hyperedge, as is defined. 
Furthermore, heterogeneous hypergraph has the property that $|\mathcal{T}_v|+|\mathcal{T}_e|>2$. 
Hence, a hyperedge $e\in\mathcal{E}$ can also be denoted as the set of all the vertices it connects: $\{v_{i_1}, v_{i_2},...,v_{i_k}\}\subseteq\mathcal{V}$.

\subsection{Hypergraph Convolutional Network (HCN)}
\label{sec::HGConv}
Graph neural networks (GNNs)~\cite{scarselli2008graph,hamilton2017inductive,wu2020comprehensive} are a family of graph-based models for learning node representations with neighborhood aggregation or message passing schema. 
\rev{Generally, GNN’s learning paradigm is based on neighborhood aggregation~\cite{gilmer2017neural}, where the node representation is learned by aggregating the embeddings of its neighbors.}
Graph convolutional network (GCN)~\cite{GCN} is one kind of typical GNN, and its crux is the layer of mean-pooling neighborhood aggregation. 
A GCN layer can be formulated as follows,
\begin{equation}
	\mathbf{h}_{v_i}^{(\ell+1)} 
	= \sigma 
	\left(
	\mathbf{W}^{(\ell+1)}\mathrm{AGG}
	(
	\{
	\mathbf{h}_{v_j}^{(\ell)} \,|\, v_j\in \mathcal{N}_{v_i}
	\}
	)
	\right),
	\label{eqn::GCN}
\end{equation}
where $\mathrm{AGG}$ denotes the aggregation function, $\mathbf{W}$ denotes the learnable weight matrix, $\sigma$ denotes non-linear activation function, $\mathbf{h}_{v_i}^{(\ell)}$ denotes the feature of node $v_i$, and $\mathcal{N}_{v_i}$ denotes the set of node $v_i$'s neighbors.

Hypergraph convolutional network (HCN)~\cite{feng2019HGNN}, as an extension of GCN, performs hypergraph convolution~(HGConv) on the hypergraph for learning latent representations.
HGConv can be viewed as a two-step dual aggregation operation~\cite{yi2020HRCN}, 
{\em node-to-edge aggregation~(n2e)} to make the hyperedge's representations contain connected vertices' features 
and {\em edge-to-node aggregation~(e2n)} to make vertex's representations contain connected hyperedges' features. 
Please note that node/edge in the graph is generalized to vertex/hyperedge in the hypergraph.
The two-step aggregation in an HCN layer can be formulated as follows, 
\begin{equation}
\begin{split}
\mathbf{f}_{e_j}^{(\ell+1)} 
& = \sigma 
\left( 
\mathbf{W}_{\text{n2e}}^{(\ell+1)}\mathrm{AGG}_{\text{n2e}}
(
\{ \mathbf{h}_{v_k}^{(\ell)} \,|\, v_k\in e_j \}
)
\right) ,
\\
\mathbf{h}_{v_i}^{(\ell+1)} 
& = \sigma 
\left( 
\mathbf{W}_{\text{e2n}}^{(\ell+1)}\mathrm{AGG}_{\text{e2n}}
(
\{  
\mathbf{f}_{e_j}^{(\ell+1)} \,|\, e_j\in \mathcal{E}_{v_i}
\}
)
\right),
\end{split}
\label{eqn::HGC-2step}
\end{equation}
where $\mathbf{W}_{\text{n2e}}, \mathbf{W}_{\text{e2n}}$ denote learnable weight matrices, 
$\mathrm{AGG}$ denotes the aggregation function, $\mathcal{E}_{v_i}$ denotes the set of hyperedges that connected to vertex $v_i$. 
$\mathbf{h}_{v_i}^{(\ell+1)}$ denote vertex $v_i$'s feature, and $\mathbf{f}_{e_j}^{(\ell+1)}$ denotes hyperedge $e_j$'s feature.
After the above two-step aggregation, the node embeddings contain the features from multiple neighbors defined by different hyperedges,  which can well capture the high-order relations.

%


\section{The Proposed Method}\label{sec::method}
\begin{figure*}[t]
	\centering
	\includegraphics[width=0.95\textwidth]{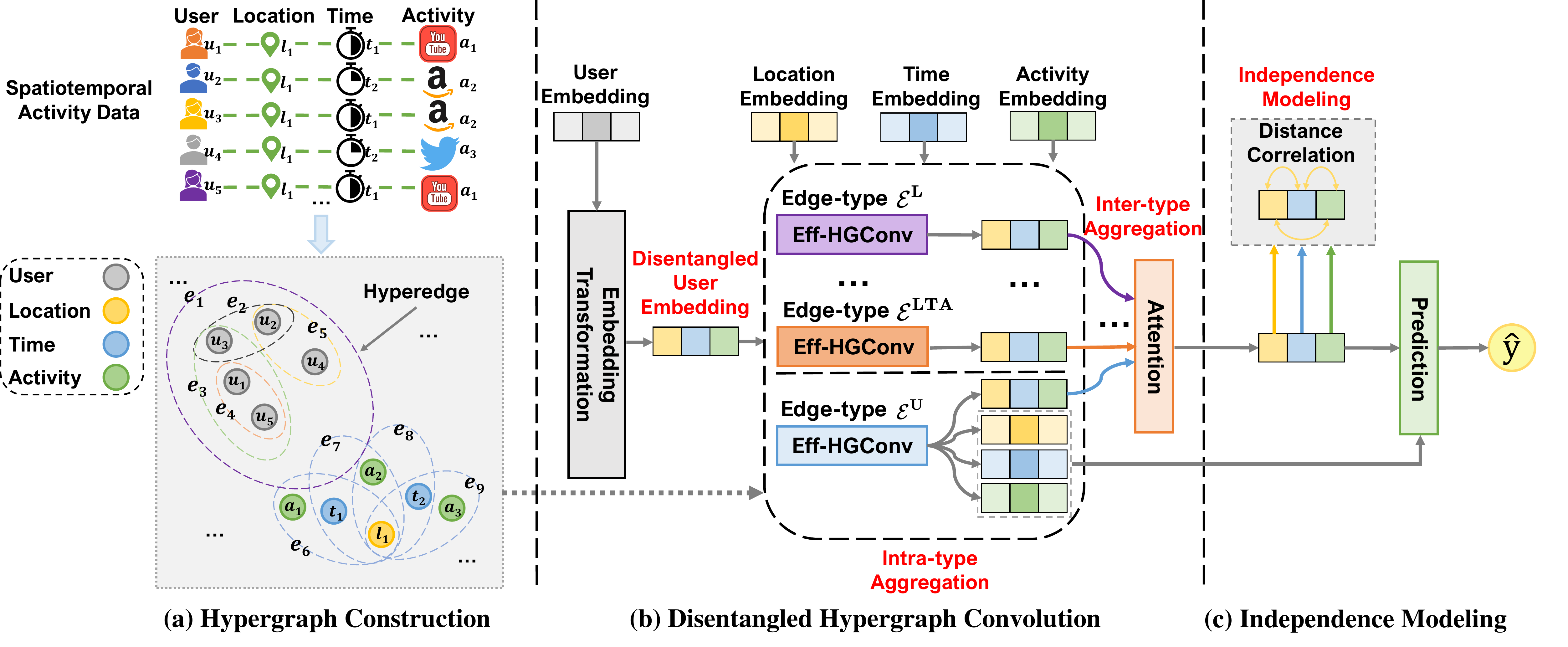}
	\setlength{\abovecaptionskip}{0.1cm}
	\setlength{\belowcaptionskip}{-0.3cm}
	\caption{Illustration of the DisenHCN model. (a) First, a heterogeneous hypergraph is constructed to model the fine-grained user similarities and the complex matching between the user and spatiotemporal activity (hyperedge colors indicate different types). (b) Then, the model projects user embeddings into three subspaces with disentangled transformation and independently conducts efficient hypergraph convolution (Eff-HGConv) to gather the high-order information in each subspace, which contains intra-type and inter-type aggregation among different types of hyperedges. We further introduce (c) independence modeling to ensure the independence of disentangled user embeddings in each subspace.
		Finally, the model predicts the probability for all activities with the given user, location, and time.
		Best viewed in color.}
	\label{fig::DisenHCN}
\end{figure*}

To overcome the two challenges in Section~\ref{sec::intro}, 
we first capture the fine-grained user similarities and the complex matching in spatiotemporal activity prediction by constructing heterogeneous hypergraph, and then we propose disentangled hypergraph convolutional networks for modeling the multi-aspect user preference and match it with spatiotemporal activity.
Fig.~\ref{fig::DisenHCN} illustrates our proposed DisenHCN model, which contains the following three parts.
\begin{itemize}[leftmargin=*]
	\item \textbf{Heterogeneous Hypergraph Construction.} 
	By converting the USTA data into a heterogeneous hypergraph with multi-type nodes and hyperedges, 
	we explicitly represent the
	fine-grained user similarities. 


\item \textbf{Disentangled Hypergraph Convolution Layer.} 
By projecting user embeddings into three aspects of preferences (\textit{i.e.} location-aware, time-aware, and activity-aware), 
we disentangle the impact of each aspect in the user's final decision-making.
With the distinct types of hyperedges in a heterogeneous hypergraph, 
we can distinguish user similarities and the matching of \textit{user-centered} interactions in different aspects.
The disentangled efficient hypergraph convolution conducts intra-type and inter-type aggregation among different types of hyperedges in each aspect independently, 
which fuses the semantic information behind user similarities and the complex matching between the user and spatiotemporal activity.

\item \textbf{Prediction Layer with Independence Modeling Loss.} 
We calculate the preference scores from three aspects (location, time, and activity) with the disentangled user representations and then fuse them to obtain the final prediction results. To make sure the disentangled user embeddings can independently capture the three aspects, we propose an independence-modeling loss, which can be jointly optimized along with the prediction loss. 
\end{itemize}

In the following, we elaborate on the three parts in detail \rev{and we explain the notations used in this paper in Table~\ref{tab::symbal}.}
\subsection{Heterogeneous Hypergraph Construction}\label{sec::hyper_construct}

Formally, let $\mathcal{G}=({\mathcal V},{\mathcal E}, \mathcal{T}_v, \mathcal{T}_e)$ be the constructed heterogeneous hypergraph, where ${\mathcal V},{\mathcal E}, \mathcal{T}_v, \mathcal{T}_e$ denote the vertex set, hyperedge set, vertex-type set and hyperedge-type set, respectively. Here $\mathcal{V}=\mathcal{U}\cup\mathcal{L}\cup\mathcal{T}\cup\mathcal{A}$ and $\mathcal{T}_v=\{\mathcal{U},\mathcal{L},\mathcal{T}, \mathcal{A}\}$, where $\mathcal{U}, \mathcal{L}, \mathcal{T}, \mathcal{A}$ denote the set of all users, locations, time slots and activities in observed USTA data $\mathcal{D}$. 

With these vertices, the hyperedges can be constructed from two perspectives, user similarity, and one-to-many matching, as follows.

\subsubsection{Hyperedges connecting users with multiple granularities}
As emphasized above, 
user similarities have multiple types (users who ``\textit{performed same activity}" or ``\textit{on the same location}" or ``\textit{at the same time}" indicates they have a similar preference of activity, location, or time, respectively). 
Meanwhile, user similarities are multi-grained 
due to the various combinations of multi-type similarities. 

To well handle the multi-type and multi-grained user similarities,
we construct seven types of hyperedges as
$
\mathcal{T}_e^u=\{\mathcal{E}^{\text{L}},
\mathcal{E}^{\text{T}},
\mathcal{E}^{\text{A}},
\mathcal{E}^{\text{LT}},
\mathcal{E}^{\text{LA}},
\mathcal{E}^{\text{TA}},
\mathcal{E}^{\text{LTA}}\}.$
Specifically,
$\mathcal{E}^{\text{L}}$ denotes the hyperedges that connect users who ``\textit{visited the same location}". For example, as Figure~\ref{fig::DisenHCN} (a) shows, users $u_1, u_2, u_3, u_4, u_5$ visited the same location $l_1$ and a corresponding hyperedge $e_1=\{u_1, u_2, u_3, u_4, u_5\}\in \mathcal{E}^{\text{L}}$ is constructed to model the preference similarity on location. 
Similarly, $\mathcal{E}^{\text{T}}$ and $\mathcal{E}^{\text{A}}$ are constructed to capture preference similarity on time and activity respectively.
With a finer granularity, 
$\mathcal{E}^{\text{LT}}$ connect users who ``\textit{visited same location at the same time}" (preference similarity on both location and time).
Then $\mathcal{E}^{\text{LA}}$ and $\mathcal{E}^{\text{TA}}$ are similarly constructed.
The last hyperedge  $\mathcal{E}^{\text{LTA}}$ has the finest granularity, which connects users who ``\textit{performed same activity on the same location at the same time}".


%
%
%

\subsubsection{Hyperedges connecting user and spatio-temporal activity}
Spatiotemporal activity prediction is essentially a complex matching problem between the user and spatiotemporal activity. In other words, the user interacts with location, time, and activity simultaneously, 
which is {\em user-centered} one-to-many interaction. 
Therefore we construct hyperedges $\mathcal{E}^{\text{U}}$, connecting all the locations, times and activities in each user's historical records.
For example, given the record of $u_1$ in Fig.~\ref{fig::DisenHCN}, 
we construct a hyperedge $e_6=\{l_1, t_1, a_1\}\in \mathcal{E}^{\text{U}}$ to model the complex interactions between $u_1$ and the locations, times, activities (\textit{i.e.} $l_1, t_1, a_1$).
That is, each hyperedge in $\mathcal{E}^{\text{U}}$ corresponds to one user.

\vspace{0.2cm}
To summarize it, we obtain hyperedge sets as follows
,
\begin{equation}
	\begin{aligned}
		\mathcal{T}_e=\{\mathcal{E}^{\text{L}},
		\mathcal{E}^{\text{T}},
		\mathcal{E}^{\text{A}},
		\mathcal{E}^{\text{LT}},
		\mathcal{E}^{\text{LA}},
		\mathcal{E}^{\text{TA}},
		\mathcal{E}^{\text{LTA}},
		\rev{\mathcal{E}^{\text{U}}}
		\}.
	\end{aligned}
\end{equation}



\subsection{Disentangled Hypergraph Convolution Layer}


\subsubsection{Disentangled Embedding Transformation}\label{sec::disen_trans}
\rev{In spatiotemporal acticity prediction, user behaviors are driven by three aspects of preferences (\textit{i.e.}, location-aware, time-aware and activity-aware) simultaneously. For example, a user $u$ performed activity $a$ on location $l$ at time $t$. In this event, the final behavior of user $u$ is determined by his/her preferences on location $l$, time $t$, or activity $a$. Obviously, disentangling those three aspects of preferences will enhance model capability and contribute to better performance.} 
To disentangle the embeddings for modeling user's preferences on multiple aspects, \textit{i.e.}, location, time, and activity,
we separate the user embeddings into three chunks.
Each chunk is associated with one aspect of user preferences (\textit{i.e.} location-aware, time-aware and activity-aware aspect).
Specifically, for each user $u\in\mathcal{U}$, we slice its raw feature vector $\mathbf{p}_u\in\mathbb{R}^d$ into three different subspaces as follows,
\begin{equation}
	\mathbf{p}_{u} = (\mathbf{p}_{u,L}, \mathbf{p}_{u,T}, \mathbf{p}_{u,A})
	\label{eqn::user-disen}
\end{equation}
where $s \in \{L, T, A\}$ represents one certain aspect of the set of all disentangled ones,
and $d$ denotes \rev{the} embedding size.
$\mathbf{p}_{u,L}\in \mathbb{R}^{\frac{d}{3}}, \mathbf{p}_{u,T}\in \mathbb{R}^{\frac{d}{3}}, \mathbf{p}_{u,A}\in \mathbb{R}^{\frac{d}{3}}$ denote the disentangled embedding chunk of user $u$ in the aspect of location, time, and activity, respectively.

As for the embeddings of locations, time and activities, 
\rev{they} are naturally embedded into the corresponding subspace.
We use
$\mathbf{q}_{l}, \mathbf{r}_{t}, \mathbf{s}_{a}\in\mathbb{R}^{\frac{d}{3}}$ to denote the initialized embedding of location $l$, time $t$ and activity $a$, which are in the subspace $L, T, A$, respectively. 
Note that each disentangled chunk of user embeddings and the embeddings of corresponding aspect  (\textit{i.e.} $\mathbf{p}_{u,L}$ and $\mathbf{q}_l$ ) are in the same subspace ($L$).

With the disentangled embedding layers, we obtain separated user embeddings on different subspaces that encode users' preferences of three different aspects. 
Here we propose the efficient hypergraph convolutions, including
the intra-type aggregation that propagate and aggregate embeddings through one type of hyperedge 
and the inter-type aggregation that involves multiple types of hyperedges.

\subsubsection{Intra-type Aggregation with Eff-HGConv}\label{sec::intra-type}

We first propose an efficient hypergraph convolution (Eff-HGConv) operation that can be used in intra-type aggregation,
which follows the two-step aggregation (including {\em node-to-edge aggregation~(n2e)} and {\em edge-to-node aggregation~(e2n)}). 

Distinct from the standard hypergraph convolution that defined in~\eqref{eqn::HGC-2step},
following~\cite{he2020lightgcn}, we remove the feature
transformation and nonlinear activation, and propose an efficient hypergraph convolution (Eff-HGConv).
Specifically, the Eff-HGConv layer can be formulated as follows, 
\begin{equation}
	\begin{split}
		\mathbf{f}_{e_j}^{(\ell+1)} & =
		\mathrm{AGG}_{\text{n2e}}
        (
        \{ \mathbf{h}_{v_k}^{(\ell)} \,|\, v_k\in e_j \}
        )
		\\
		\mathbf{h}_{v_i}^{(\ell+1)} & =
		\mathrm{AGG}_{\text{e2n}}
        (
        \{  
        \mathbf{f}_{e_j}^{(\ell+1)} \,|\, e_j\in \mathcal{E}_{v_i}
        \}
        ),
	\end{split}
	\label{eqn::Eff-HGConv}
\end{equation}
where 
$\mathbf{h}_{v_i}^{(\ell+1)}$ and $\mathbf{f}_{e_j}^{(\ell+1)}$ denote the feature of vertex $v_i$ and hyperedge $e_j$, respectively. $\mathcal{E}_{v_i}$ denotes the set of hyperedges that contain vertex $v_i$.
Here, we use the mean function as aggregation function in $\mathrm{AGG}_{\text{n2e}}$ and $\mathrm{AGG}_{\text{e2n}}$, which is demonstrated a simple yet effective choice. 
For convenience, we simplify the formulation of Eff-HGConv in \eqref{eqn::Eff-HGConv} with a single symbol $F$ as,
\begin{equation}
	[ 
	\mathbf{h}_{v_i}^{(\ell+1)}, \mathbf{f}_{e_j}^{(\ell+1)}
	]  
	= F
	( 
	\{
	\mathbf{h}_{v_k}^{(\ell)} | v_k\in e_j, e_j\in \mathcal{E}_{v_i}
	\}
	) ,
	\label{eqn::f_Eff-HGConv}
\end{equation}
which conduct feature propagation of vertex $v_i$ and its connected hyperedges \rev{$e_j\in \mathcal{E}_{v_i}$} to capture the high-order relations in hypergraph.

In the following, we introduce how to adopt Eff-HGConv to our hypergraph and hyperedges.
The intra-type aggregation with each type of hyperedge correspond to each disentangled subspace, which can applied to both 1) hyperedges connecting users with fine granularities. and 2) hyperedges connecting user and spatiotemporal activity.

\noindent \textit{i. Hyperedges connecting users with fine granularities.} 
	As mentioned in hypergraph construction, we construct hyperedges 
	$\mathcal{T}_e^u=\{\mathcal{E}^{\text{L}}, \mathcal{E}^{\text{T}}, \mathcal{E}^{\text{A}}, \mathcal{E}^{\text{LT}}, \mathcal{E}^{\text{LA}}, \mathcal{E}^{\text{TA}}, \mathcal{E}^{\text{LTA}}\}$ to model the user similarities. In this stage, for each hyperedge type $\mathcal{E}^{t_e}$, we conduct the embedding propagation on the corresponding subpace, $\mathcal{S}_{\mathcal{E}^{t_e}}$, through $\mathcal{E}^{t_e}$-type hyperedges\footnote{For each hyperedge type $\mathcal{E}^{t_e}\in\mathcal{T}_e^u$, we use $\mathcal{S}_{\mathcal{E}^{t_e}}\subseteq\mathcal{S}$ to denote the corresponding aspect (subspace) set of $\mathcal{E}^{t_e}$. For example, $\mathcal{E}^{\text{L}}$ models the complex relations among users who have similar location preference and its aspect set $\mathcal{S}_{\mathcal{E}^{\text{L}}}=\{L\}$. Hence, we only propagate the location-aware aspect of user embeddings through $\mathcal{E}^{\text{L}}$.}. 
	Then the intra-aggregation in disentangled subspace among each type of hyperedges $\mathcal{E}^{t_e}\in\mathcal{T}_e^u$ can be formulated as follows,
	\begin{equation}
	\mathbf{p}_{u_i,s,t_e}^{(\ell+1)}
	\!\! = \!
	\begin{cases}
		F(\{\mathbf{p}_{u_k,s,t_e}^{(\ell)}|u_k\in e_j, e_j\in \mathcal{E}_{v_i}^{t_e}\})
		\!\! & \! s \! \in \mathcal{S}_{\mathcal{E}^{t_e}},\\
		\mathbf{p}_{u_i,s,t_e}^{(\ell)}
		\!\! & \! s \! \notin \mathcal{S}_{\mathcal{E}^{t_e}},
	\end{cases}
	\!\!
	\label{eqn::Intra-u}
\end{equation}
	where $\mathcal{E}_{v_i}^{t_e}$ denotes the hyperedge set connected to vertex $v_i$ on hyperedge type $\mathcal{E}^{t_e}$,
	$\mathbf{p}_{u_i,s,t_e}^{(\ell)}\in \mathbb{R}^{\frac{d}{3}}$ denotes the disentangled chunk of user $u_i$ in subspace $s$ on hyperedge type $\mathcal{E}^{t_e}$ at $\ell$-th layer and $\mathbf{p}_{u_i,s}^{(0)}=\mathbf{p}_{u_i,s}$, $F$ denotes our proposed Eff-HGConv operation in~\eqref{eqn::f_Eff-HGConv}. 
	
	Note that we ignore the hyperedge feature and update the vertex feature in~\eqref{eqn::Intra-u} to obtain type-specific user representations in each disentangled subspace.
	
\noindent\textit{ii. Hyperedges connecting user and spatiotemporal activity.}
	The last type of hyperedge $\mathcal{E}^{\text{U}}$ connects all the locations, times and activities in each user's historical records, to model the one-to-many interaction for each user. 
	Hence, the complex matching between user and spatiotemporal activity can be naturally obtained through the \textit{node-hyperedge-node} aggregation with Eff-HGConv layer.
To disentangle the impact of each subspace,
	we conduct information propagation in each disentangled aspect independently. Then the intra-aggregation with $\mathcal{E}^{\text{U}}$-type hyperedge can be formulated as follows,
		\begin{equation}
\begin{split}
[\mathbf{q}_{l_i}^{(\ell+1)}, \mathbf{p}_{u,L,U}^{(\ell+1)}] &= F(\{\mathbf{q}_{l_m}^{(\ell)} |l_m\in e_u, e_u\in \mathcal{E}_{l_i}^U\}), \\
[\mathbf{r}_{t_j}^{(\ell+1)}, \mathbf{p}_{u,T,U}^{(\ell+1)}] &= F(\{\mathbf{r}_{t_n}^{(\ell)} |t_n\in e_u, e_u\in \mathcal{E}_{t_j}^U\}), \\
[\mathbf{s}_{a_k}^{(\ell+1)}, \mathbf{p}_{u,A,U}^{(\ell+1)}] &= F(\{\mathbf{s}_{a_o}^{(\ell)} |a_o\in e_u, e_u\in \mathcal{E}_{a_k}^U\}),
\label{eqn::Intra-lta}
\end{split}
\end{equation}
	where $\mathbf{q}_{l_i}^{(\ell+1)}$, $\mathbf{r}_{t_j}^{(\ell+1)}$, and $\mathbf{s}_{a_k}^{(\ell+1)}$ denote the embedding of location $l_i$, time $t_j$ and activity $a_k$, and we have $\mathbf{q}_{l_i}^{(0)}=\mathbf{q}_{l_i}, \mathbf{r}_{t_j}^{(0)}=\mathbf{r}_{t_j}, \mathbf{s}_{a_k}^{(0)}=\mathbf{s}_{a_k}$.
	Here $\mathcal{E}_{l_i}^U$, $\mathcal{E}_{t_j}^U$, and $\mathcal{E}_{a_k}^U$ denote the hyperedge set connected to location $l_i$, time $t_j$ and activity $a_k$ on hyperedge type $\mathcal{E}^{U}$, respectively, $e_u$ denotes the hyperedge for user $u$,
	and $\mathbf{p}_{u,s,U}^{(\ell+1)}$ denotes hyperedge $e_u$'s feature in subspace $s\in\mathcal{S}$
	 on hyperedge type $\mathcal{E}^{\text{U}}$
	, 
	which captures the complex matching between user and spatiotemporal activity in each disentangled subspace.

\subsubsection{Inter-type Aggregation} \label{sec::inter-type}
With the intra-type aggregation, we can obtain the type-specific user embeddings in each disentangled subspace.
To obtain a more comprehensive user embedding that model the cross-type relations, we conduct inter-type aggregation among all hyperedge types.
 \rev{Please note that the type-specific disentangled user chunks, which correspond to different types of hyperedges, have distinct impacts on the final user representation. Therefore,}
 for each user $u$, we fuse the type-specific disentangled chunks from hyperedge type in $\mathcal{T}_e$ \rev{with the attention mechanism~\cite{vaswani2017attention}, denoted as follows,}
\begin{equation}
\begin{split}
\alpha_{u,s,t_e} 
&= \frac{\mathrm{exp}(\mathbf{a}_s^\top\mathrm{tanh}(\mathbf{W}_s^{\text{int}}\mathbf{p}_{u,s, t_e}^{(\ell)}+\mathbf{b}_s^{\text{int}}))}
{\sum_{\mathcal{E}^{t_e'} \in \mathcal{T}_e}{\mathrm{exp}(\mathbf{a}_s^\top\mathrm{tanh}(\mathbf{W}_s^{\text{int}}\mathbf{p}_{u,s, t_e'}^{(\ell)}+\mathbf{b}_s^{\text{int}}))}}, 
\\
\mathbf{p}_{u,s}^{(\ell)} &= \sum_{\mathcal{E}^{t_e} \in \mathcal{T}_e} \alpha_{u,s, t_e} \mathbf{p}_{u,s,t_e}^{(\ell)},
\end{split}
\label{eqn::Inter-agg}
\end{equation}
where $\mathbf{W}_s^{\text{int}} \in \mathbb{R}^{\frac{d}{3}\times \frac{d}{3}}$, $\mathbf{b}_s^{\text{int}} \in \mathbb{R}^{\frac{d}{3}}$, and $\mathbf{a}_s^{\text{int}} \in \mathbb{R}^{\frac{d}{3}}$ denote the weight matrix, bias vector and attention vector in subspace $s\in\mathcal{S}$, respectively. 
Here $\mathbf{p}_{u,s, t_e}^{(\ell)}$ denote the disentangled chunk of user $u$ in subspace $s$ on hyperedge type $\mathcal{E}^{t_e}$ at $\ell$-th layer.
With the learned coefficients $\alpha_{u,c,t_e}$, we vary the impact from different hyperedge-types and subspaces.

\subsubsection{Output Layer}
After conducting the embedding aggregation by $L$ times, 
we obtain the $L$ embeddings from different layers, which can be further combined.
Specifically,
we combine them to obtain the final embeddings as follows, 
\begin{equation}
	\begin{split}
		\bar{\mathbf{p}}_{u,s}=\frac{1}{L+1}\sum_{\ell=0}^L\mathbf{p}_{u,s}^{(\ell)},
		\quad
		&\bar{\mathbf{q}}_l=\frac{1}{L+1}\sum_{\ell=0}^L\mathbf{q}_l^{(\ell)},
		\\
		\bar{\mathbf{r}}_t=\frac{1}{L+1}\sum_{\ell=0}^L\mathbf{r}_{t}^{(\ell)},
		\quad
		&\bar{\mathbf{s}}_a=\frac{1}{L+1}\sum_{\ell=0}^L\mathbf{s}_{a}^{(\ell)},
	\end{split}
	\label{eqn::combine}
\end{equation}
where $\bar{\mathbf{p}}_{u,s}$ denotes the disentangled feature of user $u$ in subspace $s\in\mathcal{S}$. 
Here $\bar{\mathbf{q}}_l$, $\bar{\mathbf{r}}_t$, 
and $\bar{\mathbf{s}}_a$ denote the final embedings of location $l$, time $t$ and activity $a$, respectively. 
Finally, we conduct the inner product of each aspect to estimate each one's impact on the user behavior and sum them as the final score,
\begin{equation}
	\hat{y}_{ulta}
	= \bar{\mathbf{p}}_{u,L}^\top\bar{\mathbf{q}}_l
	+ \bar{\mathbf{p}}_{u,T}^\top\bar{\mathbf{r}}_t
	+ \bar{\mathbf{p}}_{u,A}^\top\bar{\mathbf{s}}_a.
	\label{eqn::pred}
\end{equation}
In short, we build efficient hypergraph convolution layers that includes intra/inter-type aggregations, 
which are supported by our proposed Eff-HGConv unit.

\subsection{Independence Modeling Loss}

Since the disentangled chunks of user should reflect different aspects of preferences, we design a module of independence modeling to guide the learning of the chunks. 
In particular, we adopt distance correlation~\cite{szekely2009brownian,szekely2007measuring} to measure both linear and nonlinear associations of any two disentangled chunks for each user, whose coefficient is zero if and only if these chunks are independent. Hence, we regard the distance correlation of any two chunks as an independence loss, which can be formulated as follows, 
\begin{equation}
\mathcal{L}_{\mathrm{IND}}
= \sum_{u\in\mathcal{U}}\sum_{s,s'\in\mathcal{S}, s\neq s'}\frac{\text{dCov}(\bar{\mathbf{p}}_{u,s},\bar{\mathbf{p}}_{u,s'})}
{\sqrt{\text{dVar}(\bar{\mathbf{p}}_{u,s})\cdot \text{dVar}(\bar{\mathbf{p}}_{u,s'})}},
\label{eqn::loss_ind}
\end{equation}
where $\text{dCov}(\cdot)$ denotes the distance covariance of two chunks,
and $\text{dVar}(\cdot)$ denotes the distance variance of each disentangled chunk.
We further optimize the model with Bayesian Personalized Ranking (BPR) loss~\cite{2012bpr}
which can promote an observed activity to rank higher than unobserved activity, defined as,
\begin{equation}
	\mathcal{L}_{\mathrm{BPR}}=\sum_{(u,l,t,a,a^*)\in \mathcal{O}}-\ln \sigma(\hat{y}_{ulta}-\hat{y}_{ulta^*})+\lambda \|\bf\Theta\|_2^2,
	\label{eqn::bpr}
\end{equation}
where $\mathcal{O}=\{(u,l,t,a,a^*)|(u,l,t,a)\in\mathcal{D},(u,l,t,a^*)\notin\mathcal{D}\}$ denotes the pairwise training set built with negative sampling, $\sigma(\cdot)$ denotes the sigmoid function, $\bf\Theta$ denotes the model parameters, and $\lambda$ denotes $L_2$-regularization coefficient to avoid over-fitting. Combining BPR loss and independence loss, we minimize a loss function as follows,
\begin{equation}
	\mathcal{L}=\mathcal{L}_{\mathrm{BPR}}+\gamma\mathcal{L}_{\mathrm{IND}},
	\label{eqn::total_loss}
\end{equation}
where $\gamma$ denotes the hyperparameter to control the influence of independence modeling.

\subsection{Efficient Implementation}
Given that we only obtain user (vertices) disentangled representations in~\eqref{eqn::Intra-u},
following HGNN~\cite{feng2019HGNN}, 
we ignore the hyperedge embeddings and provide the matrix form of propagation in equation~\eqref{eqn::Intra-u} as,
\begin{equation}
\mathbf{P}_{s,t_e}^{(\ell+1)} = \mathbf{D}_{t_e}^{-\frac{1}{2}}\mathbf{H}_{t_e}{\mathbf \Delta}_{t_e}^{-1}\mathbf{H}_{t_e}^\top\mathbf{D}_{t_e}^{-\frac{1}{2}}\mathbf{P}_{s,t_e}^{(\ell)}, \quad \mathcal{E}^{t_e}\in\mathcal{T}_e^u
\label{eqn::mul-HConv}
\end{equation}
where $\mathbf{P}_{s,t_e}^{(\ell)}$ denotes the disentangled user embedding matrix in subspace $s\in\{L,T,A\}$.
$\mathbf{D}_{t_e}$ and ${\mathbf \Delta}_{t_e}$~are diagonal matrices of vertex degree and hyperedge degree to re-scale the embeddings~\cite{yu2021MHCN}. $\mathbf{H}_{t_e}$ is the incidence matrix of hyperedge $\mathcal{E}^{t_e}\in\mathcal{T}_e^u$. 
Obviously, we can easily construct the incidence matrices according to Section~\ref{sec::hyper_construct}. For example, we have $\mathbf{H}_{L}\in \mathbb{R}^{N_U\times N_L}, \mathbf{H}_{T}\in \mathbb{R}^{N_U\times N_T}, \mathbf{H}_{A}\in \mathbb{R}^{N_U\times N_A}$ to represent the user similarities in aspects of location, time and activity, respectively.
However, the incidence matrix of $\mathbf{H}_{LTA}\in \mathbb{R}^{N_U\times (N_L N_T N_A)}$ have $N_L \times N_T \times N_A$ columns. Thus, it's impossible to directly build the incidence matrix $\mathbf{H}_{LTA}$ on large-scale datasets. 

Luckily, we do not need $\mathbf{H}$ and only need $\mathbf{H}{\mathbf{H}^\top}$ in convolutional operations if we ignore $\mathbf{D}_c$ and ${\mathbf \Delta}_c$ that only re-scale embeddings. Here we use $\mathbf{A}=\mathbf{H}{\mathbf{H}^\top}$ for simplification. Given that the element of incidence matrix is binary value, we can use multiplication to represent logical \texttt{AND} operator ($\&$) and obtain the logical relations among different types of hyperedges that capture user similarities.
Inspired by matrix factorization~\cite{lowrankMF}, we further propose a equivalent calculation for Eff-HGConv. Specifically, we can calculate $\mathbf{A}_{LTA}=\mathbf{H}_{LTA}{\mathbf{H}_{LTA}^\top}$ with $\mathbf{A}_{L}, \mathbf{A}_{T}, \mathbf{A}_{A}$, which is formulated as follows,
\begin{equation}
\mathbf{A}_{LTA}=\mathbf{A}_{L}\odot\mathbf{A}_{T}\odot\mathbf{A}_{A},
\label{eqn::Aii}
\end{equation}
where $\odot$ denotes the \textit{Hadamard} product. With $\mathbf{A}_{L}, \mathbf{A}_{T}, \mathbf{A}_{A}$, we can further calculate $\mathbf{A}_{LT}, \mathbf{A}_{LA}, \mathbf{A}_{TA}$ in a similar manner. With the above calculation, following~\cite{yu2021MHCN}, we can easily simplify the hypergraph convolution in~\eqref{eqn::mul-HConv} with the equivalent matrices as follows,
\begin{equation}
    \mathbf{P}_{s,t_e}^{(\ell+1)} = {\mathbf{D}^*}_{t_e}^{-\frac{1}{2}} \mathbf{A}_{t_e}{\mathbf{D}^*}_{t_e}^{-\frac{1}{2}} \bm{P}_{s,t_e}^{(\ell)},
\label{eqn::eq_conv}
\end{equation}
where ${{\mathbf{D}^*}}_{t_e}$ is the degree matrix of $\mathbf{A}_{t_e}$. With the above equivalent, we reduce the time complexity of propagation in equation~\eqref{eqn::Intra-u} from $\mathcal{O}(N_U^2N_LN_TN_AdL)$ to $\mathcal{O}(N_U^2dL)$.

\subsection{Complexity Analysis}

\subsubsection{Space Complexity} 
\rev{The trainable parameters of node embeddings in GCN only include the 0-th layer initialized embeddings~\cite{he2020lightgcn} and node embeddings in layer $1$-$L$ are calculated by propagation based on the 0-th layer embeddings.}
Hence, the trainable parameters of DisenHCN includes the embeddings of the $0$-th layer and attention parameters, where the embeddings' parameter size is far larger than other one.
The embeddings of $0$-th layer consist of user embeddings $\mathbf{P}^{(0)}\in \mathbb{R}^{N_U \times d}$, location embeddings $\mathbf{Q}^{(0)}\in \mathbb{R}^{N_L \times \frac{d}{3}}$, time embeddings $\mathbf{R}^{(0)}\in \mathbb{R}^{N_T \times \frac{d}{3}}$ and activity embeddings $\mathbf{S}^{(0)}\in \mathbb{R}^{N_A \times \frac{d}{3}}$, where $d$ is embedding size. 
Note that our DisenHCN just takes $d/3$ as the embedding dimensionality of $\mathbf{Q}, \mathbf{R}, \mathbf{S}$ but achieves better performance.
Hence, the model size of DisenHCN is lighter than the most simplified tensor models.

\subsubsection{Time Complexity} 
\rev{Following~\cite{he2020lightgcn}, 
we remove the nonlinear feature transformations in each layer.}
The computation cost of DisenHCN mainly derives from hypergraph convolution and attention operation. 
In the hypergraph convolution on heterogeneous hypergraph, 
the time cost is $\mathcal{O}(4N_U^2dL+(N_L+N_T+N_A)N_UdL/3)$, where $L$ denotes the number of hypergraph convolution layers.
Moreover, attention operation in inter-type aggregation's time cost it $\mathcal{O}(8N_Ud^2/3)$. 
\rev{Hence, the time complexity of DisenHCN is $\mathcal{O}(4N_U^2 dL+(N_L+N_T+N_A)N_U dL/3+8N_Ud^2/3)$. However, for the existing GNN models, such as SA-GCN~\cite{yu2020SAGCN}, its time cost is $\mathcal{O}((N_U+N_L+N_T+N_A)^2d^3L)$. Given that $N_U,N_L,N_T,N_A>>d,L$ ($d=120$, $L=1,2,3,4$), the time complexity of DisenHCN is lower than most existing GNN models.} 

%
%
%
\section{Experiment}\label{sec::experiments}

\subsection{Experimental Setup}
\subsubsection{Datasets}~Our experiments are based on four real-world datasets: Telecom, TalkingData, 4SQ and TWEET. The statistics of datasets is shown in Table~\ref{tab:dataset}.
\begin{itemize}[leftmargin=*]
	\item {\bf Telecom}~\cite{Telecom} is a large-scale mobile App usage dataset collected during April 20-26, 2016 from a major mobile network operator in China. Following SA-GCN~\cite{yu2020SAGCN}, we divide the location area into small squares of size $3$km$\times$3km to generate location ID. Moreover, we distinguish weekdays and weekends and discretize a day into 24 units, generating 48 time slots.
	\item {\bf TalkingData\footnote{\url{https://www.kaggle.com/c/talkingdata-mobile-user-demographics}}} is an open dataset released in Kaggle, which contains App usage behavior records and spatiotemporal information. The pre-processing is similar to Telecom.
	\item {\bf4SQ}~\cite{crossmapwww17} is collected from Foursquare, which contains the geo-tagged check-ins posted in New York from August 2010 to October 2011. 
	Following CrossMap~\cite{crossmapwww17}, we use the hotspot detection methods to generate the spatiotemporal hotspots (to obtain location and timeslot-ID) and keywords. Note that we regard the keywords as activities, similar to CrossMap~\cite{crossmapwww17}.
	\item {\bf TWEET}~\cite{crossmapwww17} is collected from
	Twitter, which contains the geo-tagged tweets published in Los Angeles during the period from 1st August to 30th November 2014.
	The pre-processing of TWEET is similar to 4SQ.
\end{itemize}

\begin{table}[t]
	\centering
	\caption{Statistics of Datasets}
	\setlength\tabcolsep{1pt}
	\scalebox{0.9}{
		\begin{tabular}{ccccccc}
			\toprule[1pt]
			{\bf Dataset}  & \#{\bf User} & \#{\bf Location} & \#{\bf Time} & \#{\bf Activity} & \#{\bf Records} &{\bf Density}  \\
			\hline
			{\bf Telecom}       & 10,099 &2,462 &48 &1,751 &604,401 &2.9e-7     \\
			{\bf TalkingData}   & 4,068 &1,302 &48 &2,310 &409,651 &6.9e-7     \\
			{\bf 4SQ}    & 10,543 &1,012 &31 &3,960 &291,485 &2.2e-7     \\
			{\bf TWEET}   & 10,414 &2,080 &27 &5,960 &333,812 &9.6e-8     \\
			\bottomrule[1pt]
	\end{tabular}}
	\label{tab:dataset}
\end{table}

\begin{table*}[t]
	\caption{General performance comparisons on four real-world datasets.
		Improvements are compared with the best baseline.}
	\label{tab::performance}
	\centering
	\setlength\tabcolsep{2pt}
	\begin{tabular}{c|c|cc|cc|cc|cc}
		\toprule[1pt]
		\multicolumn{2}{c|}{\bf Dataset} & \multicolumn{2}{c|}{\bf Telecom} & \multicolumn{2}{c|}{\bf TalkingData} & \multicolumn{2}{c|}{\bf 4SQ} & \multicolumn{2}{c}{\bf TWEET} \\ 
		\hline
		\bf Category &\bf Method   &    \bf Recall@10        & \bf NDCG@10  &    \bf Recall@10        & \bf NDCG@10  &    \bf Recall@10        & \bf NDCG@10  & \bf Recall@10        & \bf NDCG@10 \\ \hline
		\multirow{2}{*}{Tensor}
		&SCP     &     0.3038     & 0.1874    & 0.3290  &   0.2348      & 0.1783       & 0.1259  & 0.0583   & 0.0367 \\  
		&MCTF        &     0.3257       & 0.2042    & 0.3622  &  0.2388   &   0.1875      & 0.1347    & 0.0754  &     0.0465 \\  
		\hline
		\multirow{4}{*}{Graph}    
		&CrossMap     &      0.4164      & 0.2574       & 0.4126  &     0.2652  & 0.2456       & 0.1695       & 0.0822  &         0.0503 \\  
		&SA-GCN         & 0.4289       & 0.2674       & 0.4556  &         0.2774      & 0.2583       & 0.1764      & 0.0804  &         0.0490 \\  
		&LightGCN         &      0.4687       & 0.3057       & 0.5367  &         0.3624       &   0.2887      & 0.1945      & 0.0962  &     0.0601 \\  
		&HAN     &      0.4819      & 0.3160       & 0.5517  &     0.3756    & 0.3042  & 0.2034       & 0.1005  &         0.0647 \\  
		\hline
		\multirow{4}{*}{Hypergraph} 
		&DHCF             & 0.4768       & 0.3105 & 0.5443  & 0.3744           & 0.2967     & 0.1987  & 0.0971  & 0.0603 \\  
		&MHCN             & 0.4893       & 0.3172 & 0.5596  & 0.3804  & 0.3065      & \underline{0.2094}  & 0.1014  & \underline{0.0665} \\  
		&HWNN     &      \underline{0.4958}      & \underline{0.3207}       & \underline{0.5623}  &     \underline{0.3812}    & \underline{0.3086}  & 0.2067       & \underline{0.1026}  &         0.0654 \\ 
		&\bf DisenHCN         &     \bf0.5664       & \bf0.3731     & \bf 0.6427     & \bf0.4322      &     \bf0.3576     & \bf0.2473  &  \bf0.1172    & \bf0.0771 \\  
		\hline
		\multicolumn{2}{c|}{Improvement in \%}  &   14.24\%       & 16.34\%       & 14.30\%  &   13.38\%      &      15.88\%       & 18.10\%       & 14.23\%  &     15.94\%     \\
		\bottomrule[1pt]
	\end{tabular}
\end{table*}

\subsubsection{Evaluation Metrics} Following existing works~\cite{zheng2010UCLAF,MCTF_WWW2015,fan2019personalized}, we adopt two widely-used metrics, Recall@K and NDCG@K, to evaluate the performance. 
In our experiments, 
we use the popular setting of $K=10$~\cite{he2020lightgcn,yu2021MHCN,he2017NCF} for evaluation.
\begin{itemize}[leftmargin=*]
	\item \textbf{Recall@K} measures the ratio of test activities that has been successfully predicted in the top-K ranking list. 
	\item \textbf{NDCG@K} assigns higher scores to hits at the higher position in the top-K ranking list, which emphasizes that test activities should be ranked as higher as possible.
\end{itemize}

\subsubsection{Baselines} To demonstrate the effectiveness of our 
DisenHCN model, we compare it with the following competitive methods, including tensor-based model, graph-based model and hypergraph-based model.

\textit{The two compared tensor-based models are:}
\begin{itemize}[leftmargin=*]
	\item \textbf{SCP}~\cite{fan2019personalized} is widely used in spatiotemporal activity prediction, which models user, location, time, and activity with a 4D tensor and conduct tensor decomposition to obtain latent features.
	\item \textbf{MCTF}~\cite{MCTF_WWW2015} simultaneously factorizes coupled tensors and matrices shared common modes with the tensors, of which the matrices are constructed by reducing different dimensions (user, location, time, or activity).
\end{itemize}

\textit{The four compared graph-based models are:}
\begin{itemize}[leftmargin=*]
	\item \textbf{CrossMap}~\cite{crossmapwww17} is a state-of-the-art method for spatiotemporal activity prediction with cross-modal graph embeddings.
	\item \textbf{SA-GCN}~\cite{yu2020SAGCN} is a state-of-the-art method in App-usage activity prediction and adopts Graph Convolutional Network (GCN) with meta path-based objective function to combine the structure of the graph and the attribute of units, including locations, time, and Apps.
	\item \textbf{LightGCN}~\cite{he2020lightgcn} is a state-of-the-art method to learn node representations and capture user-item interaction in collaborative filtering. We adapt LightGCN to the same graph structure with SA-GCN as another competitive baseline.
	\item \textbf{HAN}~\cite{wang2019han} is a state-of-the-art method for heterogeneous graph learning. We adapt it to the graph in our task, with four types of nodes representing user, location, time, and activity and six types of edges representing co-occurrence relations, to solve our problem.
\end{itemize}

\textit{The three compared hypergraph-based models are:}
\begin{itemize}[leftmargin=*]
	\item \textbf{DHCF}~\cite{ji2020dual} is a hypergraph-based collaborative filtering method for recommendation, which constructs two homogeneous hypergraphs for user and item respectively. 
	Since it can only handle homogeneous graphs, to adapt it to our task, we construct four homogeneous hypergraphs for user, location, time and activity.
	
	\item \textbf{MHCN}~\cite{yu2021MHCN} is another recently-proposed hypergraph learning model. It constructs a multi-channel hypergraph to capture the complex social relations among users for social recommendation. To adapt it to our problem, we construct three channels for location, time, and activity.
	Following the original paper, we also construct a user-location-time-activity interaction graph to capture the pair-wise interactions among user, location, time, and activity.
	
	\item \textbf{HWNN}~\cite{sun2021HWNN} is a state-of-the-art heterogeneous hypergraph
	embedding method. We define six types meta relations, \textit{i.e.} \{UL, UT, UA, LT, LA, TA\}, and with them we construct the heterogeneous hypergraph snapshots.
\end{itemize}

\begin{table*}[ht]
	\caption{Ablation study of the key designs (w/o denotes without).}
	\label{tab::ablation}
	\centering
	\setlength\tabcolsep{1pt}
	\begin{tabular}{c|c|cc|cc|cc|cc}
		\toprule[1pt]
		\multicolumn{2}{c|}{\bf Dataset} & \multicolumn{2}{c|}{\bf Telecom} & \multicolumn{2}{c|}{\bf TalkingData} & \multicolumn{2}{c|}{\bf 4SQ} & \multicolumn{2}{c}{\bf TWEET} \\ 
		\hline
		\bf Category &\bf Method   &    \bf Recall@10        & \bf NDCG@10  &    \bf Recall@10        & \bf NDCG@10  &    \bf Recall@10        & \bf NDCG@10  & \bf Recall@10        & \bf NDCG@10 \\ \hline
		\multirow{9}{*}{\makecell[c]{Multi-type \\ Hyperedges \\(Sec.~\ref{sec::hyper_construct})}}
		& w/o $\mathcal{E}^{\text{L}}$    &     0.5409       & 0.3562    & 0.6079     
		& 0.4112   & 0.3317      &  0.2318     & 0.1106     & 0.0718 \\
		& w/o $\mathcal{E}^{\text{T}}$     &     0.5458       & 0.3594    & 0.6115     
		& 0.4137   & 0.3293      &  0.2301     & 0.1085     & 0.0709 \\
		& w/o $\mathcal{E}^{\text{A}}$     &     0.5472       & 0.3607    & 0.6236     
		& 0.4211   & 0.3348      &  0.2337     & 0.1122     & 0.0725 \\
		& w/o $\mathcal{E}^{\text{LT}}$     &     0.5557       & 0.3685    & 0.6382     
		& 0.4291   & 0.3297      &  0.2309     & 0.1113     & 0.0718 \\
		& w/o $\mathcal{E}^{\text{LA}}$     &     0.5569       & 0.3693    & 0.6351     
		& 0.4275   & 0.3328      &  0.2328     & 0.1159     & 0.0762 \\
		& w/o $\mathcal{E}^{\text{TA}}$     &     0.5548       & 0.3681    & 0.6369     
		& 0.4301   & 0.3317      &  0.2318     & 0.1138     & 0.0743 \\
		& w/o $\mathcal{E}^{\text{LTA}}$    &     0.5317       & 0.3484    & 0.5956    
		& 0.4053   & 0.3195      &  0.2242     & 0.1099     & 0.0712 \\
		& w/o $\mathcal{E}^{\text{U}}$     &     0.5207       & 0.3425    & 0.5867     
		& 0.3994   & 0.3157      &  0.2209     & 0.1074     & 0.0703 \\
		& All hyperedges     &     \bf0.5664       & \bf0.3731     & \bf 0.6427     & \bf0.4322      &     \bf0.3576     & \bf0.2473  &  \bf0.1172    & \bf0.0771 \\
		\hline
		\multirow{2}{*}{\makecell[c]{Disen-Transition \\
		(Sec.~\ref{sec::disen_trans})}}
		& w/o Transition     &     0.5483       & 0.3576     & 0.6248     & 0.4211      &  0.3503     & 0.2432  & 0.1118     & 0.0727 \\
		& w Transition     &     \bf0.5664       & \bf0.3731     & \bf 0.6427     & \bf0.4322      &     \bf0.3576     & \bf0.2473  &  \bf0.1172    & \bf0.0771 \\
		\hline
		\multirow{2}{*}{\makecell[c]{Intra-type Agg. \\(Sec.~\ref{sec::intra-type})}}
		& HGConv    &     0.5567       & 0.3655     & 0.6369     & 0.4295      &  0.3498     & 0.2422  & 0.1128     & 0.0733 \\
		& Eff-HGConv     &     \bf0.5664       & \bf0.3731     & \bf 0.6427     & \bf0.4322      &     \bf0.3576     & \bf0.2473  &  \bf0.1172    & \bf0.0771 \\
		\hline
		\multirow{3}{*}{\makecell[c]{Inter-type Agg.\\(Sec.~\ref{sec::inter-type})}}
		& Max     &     0.5269      & 0.3452     & 0.5751     & 0.3932     &  0.3152     & 0.2209  & 0.1043     & 0.0678 \\
		& Mean    &     0.5543       & 0.3622     & 0.6359     & 0.4296      &  0.3287     & 0.2295  & 0.1131     & 0.0734 \\
		& Attention     &     \bf0.5664       & \bf0.3731     & \bf 0.6427     & \bf0.4322      &     \bf0.3576     & \bf0.2473  &  \bf0.1172    & \bf0.0771 \\
		\bottomrule[1pt]
	\end{tabular}
\end{table*}

It is worth mentioning that there exists some dynamic graph-based methods~\cite{wang2020HyperRec,yi2020HRCN,shi2020odflow,sawhney2020STHGCN} for spatiotemporal-aware problems, such as origin-destination flow~\cite{shi2020odflow} and stock movement forecasting~\cite{sawhney2020STHGCN}, which construct a spatial graph as a feature encoder for each time step and use RNN to fuse the features generated by spatial graphs. However, those dynamic graph-based methods can not be well applied to spatiotemporal activity prediction
as these methods require constructing a spatial graph for each time step, consuming huge computation resources, which has been reported in~\cite{wang2021apan} and~\cite{guo2020dynamic}. Thus, these works are always only evaluated on small-scale graph-structure data~\cite{wang2020HyperRec,yi2020HRCN,shi2020odflow,sawhney2020STHGCN}.

\subsubsection{Hyper-parameter Settings} For all the models, the embedding size of users, locations, time, and activities are fixed as 120, and the mini-batch size is set to 2048. We optimize all these models with Adam optimizer~\cite{kingma2014adam}  with an initial learning rate of  0.001,  which will decay by 0.1 after every 20 epochs. The $L_2$ regularization coefficient $\lambda$ is set as $3e^{-5}$. In terms of initialization, we adopt Xavier initializer~\cite{glorot2010Xavier} to initialize the embedding parameters.
For the baselines, the hyper-parameters are searched following the original papers to obtain optimal performance. 
For our proposed DisenHCN, we set the size of full user embedding as 120. Then the chunk size of each aspect is set to 40.
Without
specification, we fix the coefficient of independence modeling $\gamma$ as $3e^{-3}$ and the number of layers $L$ as 1.
Moreover, in Sections~\ref{sec:hyperpara}, we study impact of these hyper-parameters by varying $\gamma$ in
$\{1e^{-4}, 3e^{-4}, 1e^{-3}, 3e^{-3}, 1e^{-2}, 3e^{-2}, 1e^{-1}\}$ and $L$ in $\{1, 2, 3, 4\}$, respectively.
All the baselines and our DisenHCN are implemented in PyTorch\footnote{https://pytorch.org/}.
For all methods, we run 10 times with different random seeds and report the average results.

\subsection{Performance Comparisons}
\label{sec:4.2}

\subsubsection{Predicting Performance}
Experimental results are shown in Table~\ref{tab::performance},
where we can have the following observations.
\begin{itemize}[leftmargin=*]
	\item {\bf DisenHCN achieves the best performance.}~Our DisenHCN model significantly outperforms the state-of-the-art methods by 14.23\% to 18.10\% on all datasets. Due to the special design, DisenHCN can capture multi-type and multi-grained user similarities and complex matching between the user and spatiotemporal activity simultaneously, which contributes to better prediction performance. 
	\item {\bf Explicitly exploring the interactions among user, location, time, and activity is necessary.}~The graph-based and hypergraph-based models outperform the tensor-based methods,
	which can be explained by 
	the strong power of explicitly relation-modeling for USTA data.
	This further confirms the necessity of exploring interaction relationships of user, location, time, activity, which validates our motivation of designing hypergraph-based method.
	
	\item 
	{\bf Exploring heterogeneous information is important in the spatiotemporal activity prediction.}~
	We can observe that hypergraph baseline DHCF does not perform better than HAN, a normal graph model with heterogeneous information. It reveals that if there are no proper designs in our task, hypergraph may bring the noise, although it can theoretically model more complex relations. 
	When combining the heterogeneous information in hypergraph modeling, MHCN (with heterogeneous user similarities) and HWNN (with heterogeneous user interactions) achieve the best performance among all baselines.
	This demonstrates that our special design for hypergraph construction according to heterogeneous information (\textit{i.e.} the multiple user similarities and interactions) is necessary.
	
\end{itemize}

\rev{Note that the performance on TWEET dataset is the worst for all models.
	According to dataset statistics in Table~\ref{tab:dataset}, the relatively worse performance of TWEET dataset may be caused by the following reasons: 1) TWEET dataset is sparser; 2) the number of activities in TWEET dataset is far larger than Telecom dataset. Our DisenHCN can also achieve significant improvement on this sparse dataset, and we will further analyze the impact of data sparsity in Section~\ref{sec::sparse}.}

\begin{table}[t]
\small
	\caption{Total training time (in minutes).}
	\label{tab::training time}
	\centering
	\setlength\tabcolsep{1pt}
		\begin{tabular}{c|c|cccc}
			\toprule[1pt]
			\bf Category & \bf Method   &{\bf Telecom} & {\bf TalkingData} & {\bf 4SQ} & {\bf TWEET} \\  \hline
			\multirow{2}{*}{Tensor}
			&SCP     &     21.36     & 18.42    & 25.35  &   27.48   \\  
			&MCTF    &     21.75     & 18.68    & 25.73  &   27.84  \\  
			\hline
			\multirow{4}{*}{Graph}    
			&CrossMap      &  30.45     & 28.59    & 37.13  &   41.27  \\  
			&SA-GCN        &  29.46     & 27.32    & 35.89  &   39.67  \\  
			&LightGCN      &  28.17     & 25.93    & 34.64  &   38.28  \\  
			&HAN           &  31.07     & 29.26    & 38.25  &   42.31 \\  
			\hline
			 
			&DHCF          &     31.45     & 29.72    & 38.68  &   42.61  \\  
		Hyper-	&MHCN          &     30.79     & 29.27    & 37.95  &   41.77 \\  
		graph	&HWNN          &     32.46     & 31.47    & 40.15  &   43.94  \\ 
			&\bf DisenHCN  &     24.35     & 21.59    & 28.34  &   31.86 \\ 
			\bottomrule[1pt]
	\end{tabular}
\end{table}

\subsubsection{Efficiency Comparison}
To demonstrate the efficiency of our DisenHCN model, we report the training time until convergence in Table~\ref{tab::training time}. We conduct the
experiments on an NVIDIA TITAN-XP GPU. From the results, we can observe that, except for the classical tensor-based models, our method
is more efficient compared with existing graph/hypergraph-based methods.

\begin{figure*}[ht]
	\centering
	\subfloat{\includegraphics[width=0.48\columnwidth]{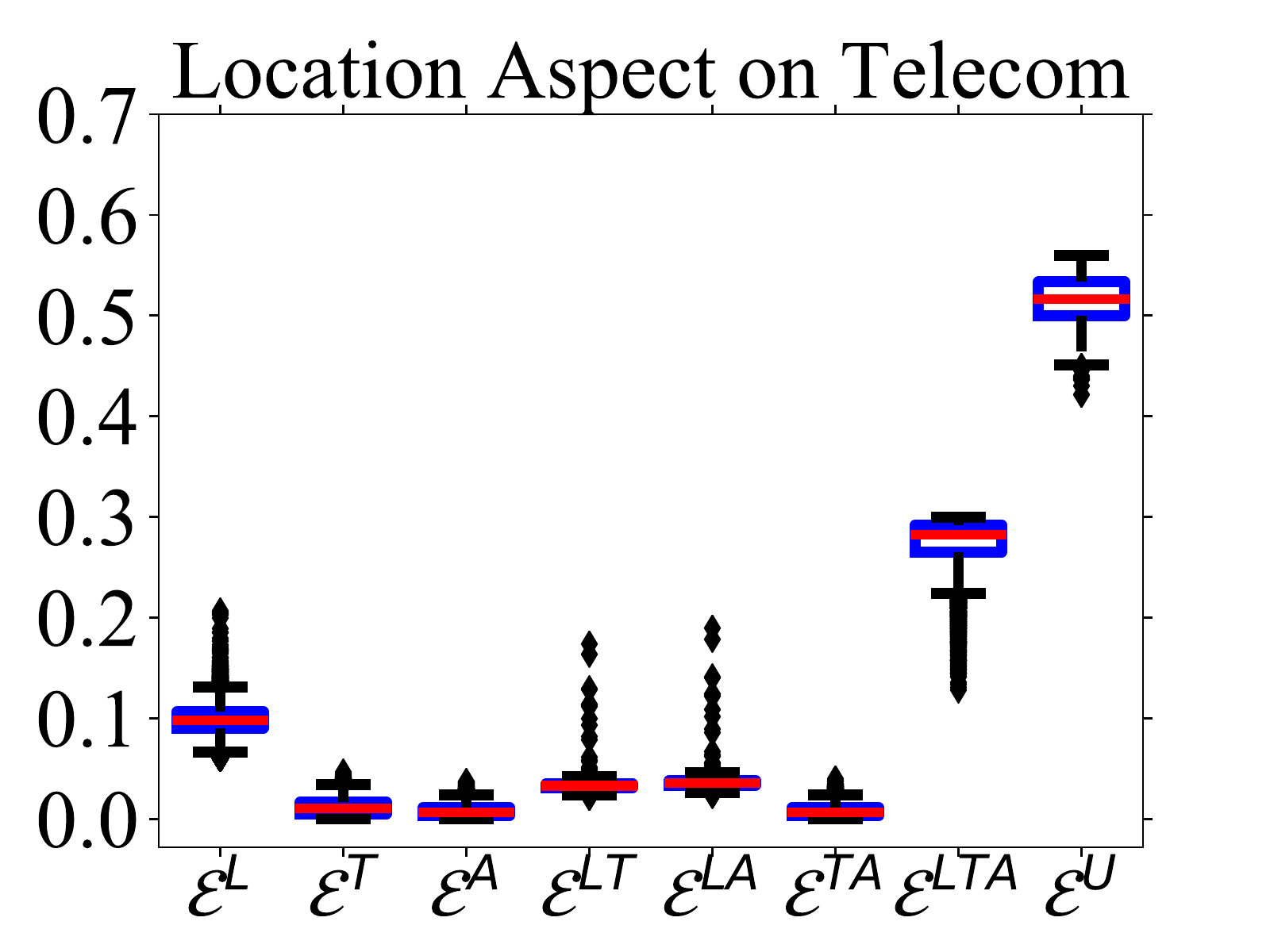}}
	\subfloat{\includegraphics[width=0.48\columnwidth]{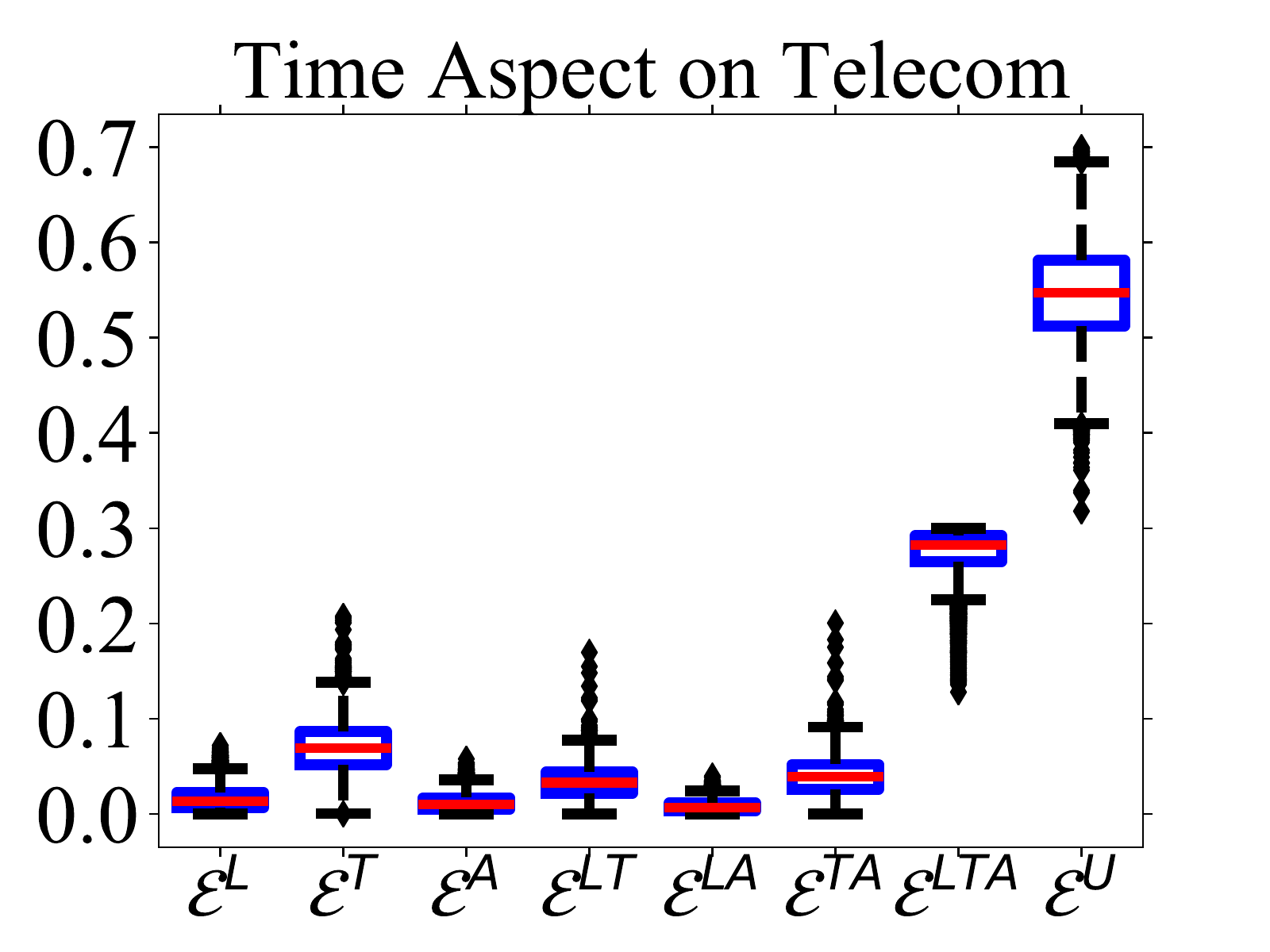}}
	\subfloat{\includegraphics[width=0.48\columnwidth]{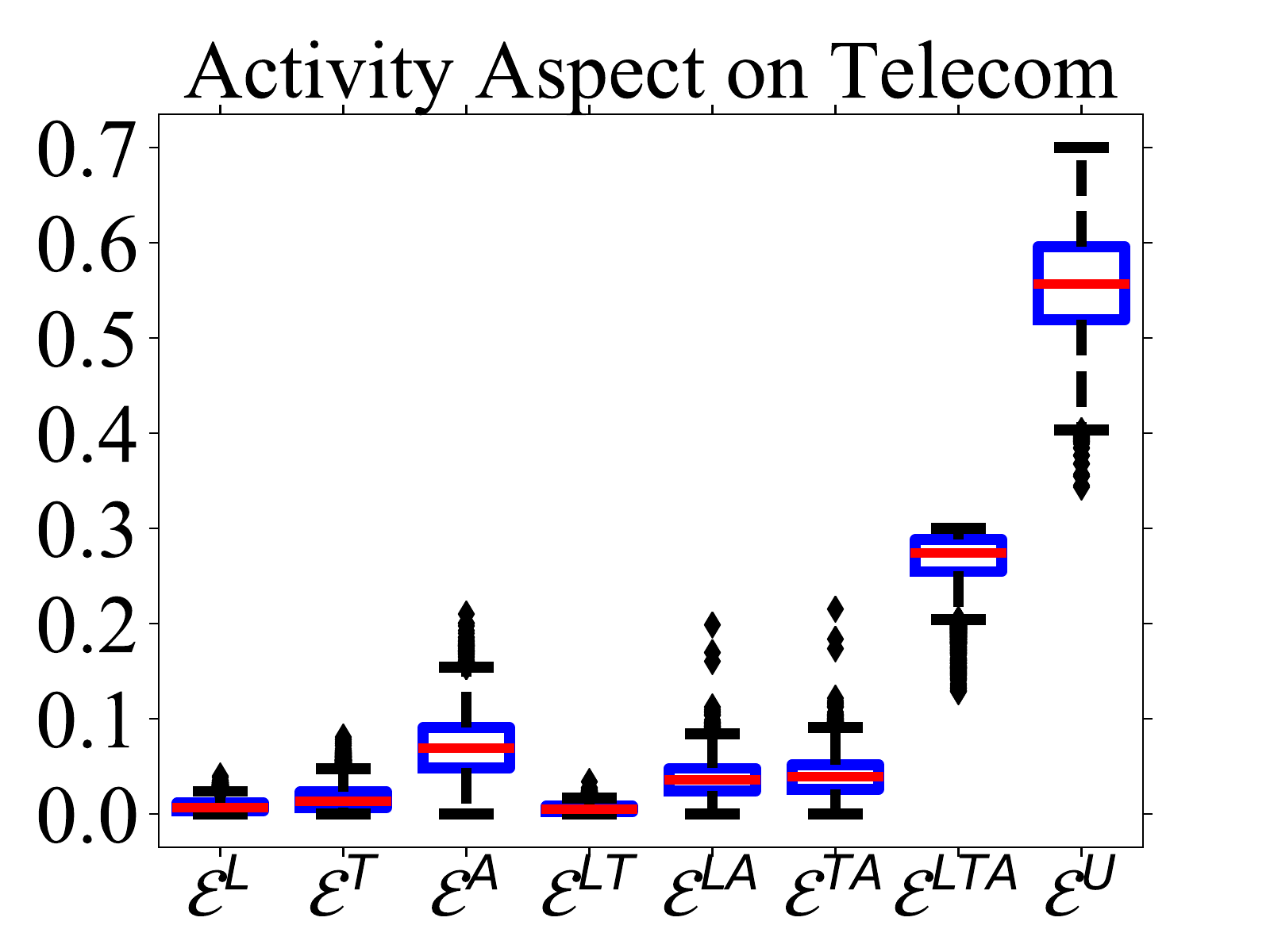}}\\
	\subfloat{\includegraphics[width=0.48\columnwidth]{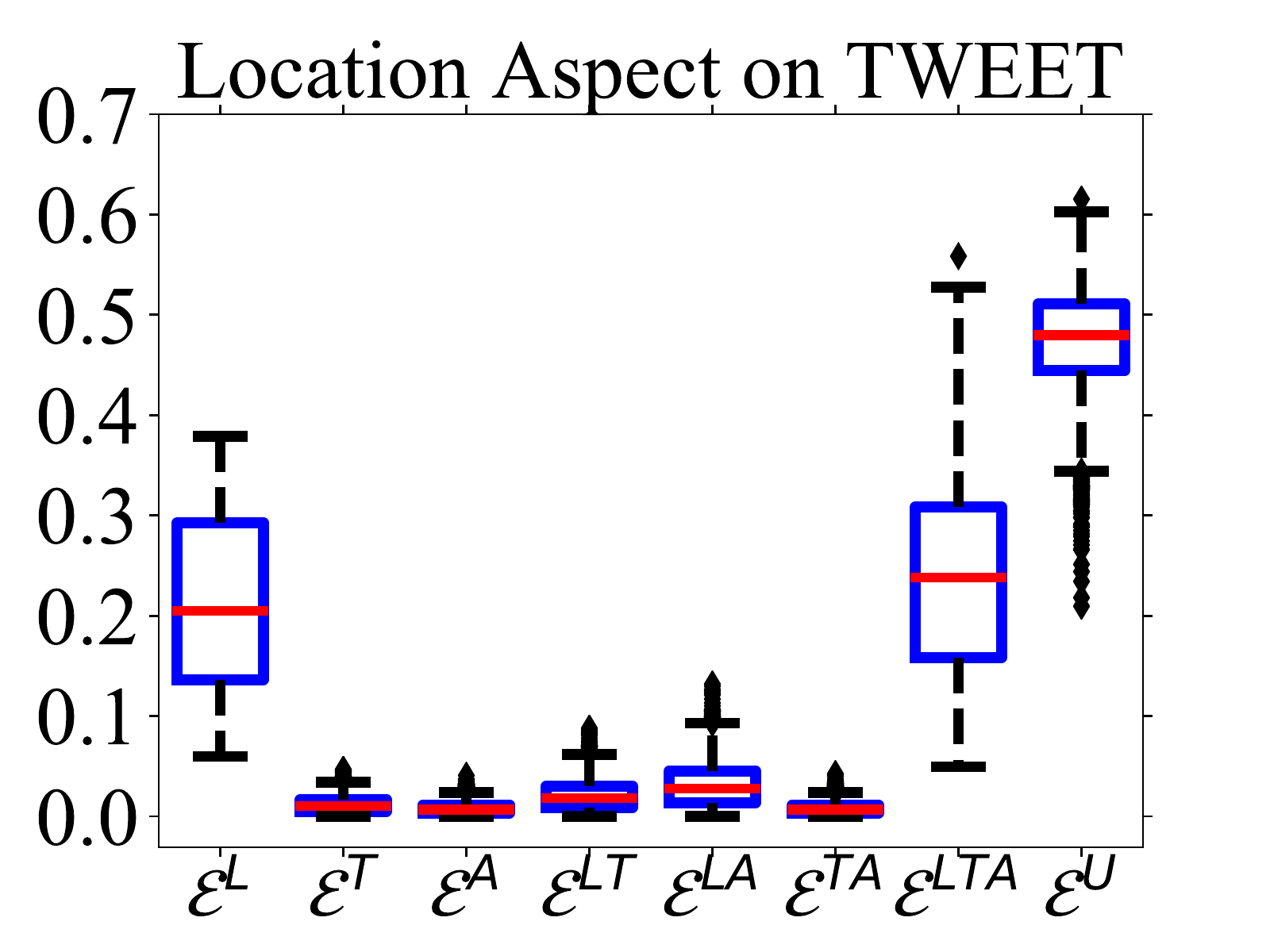}}
	\subfloat{\includegraphics[width=0.48\columnwidth]{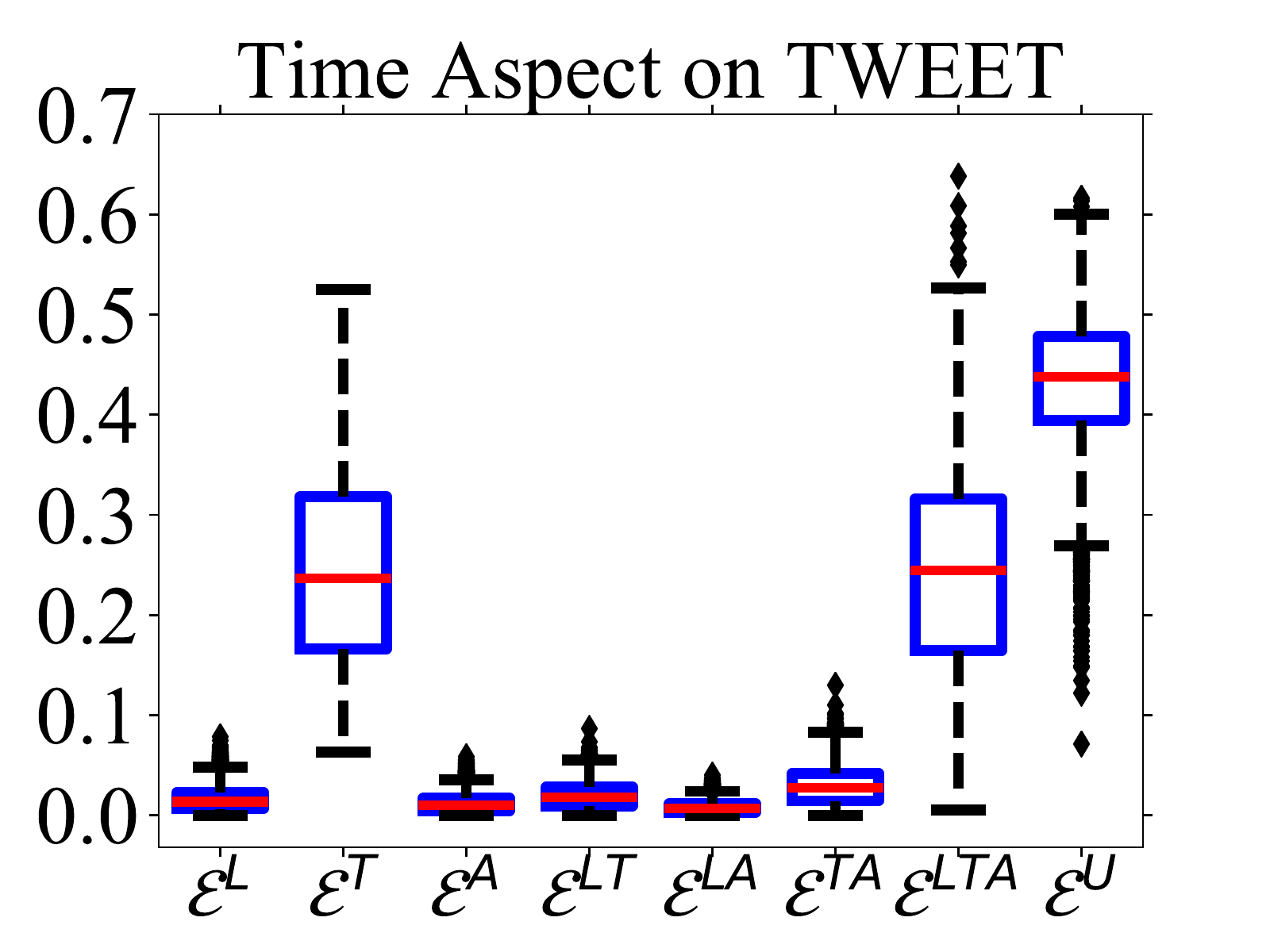}}
	\subfloat{\includegraphics[width=0.48\columnwidth]{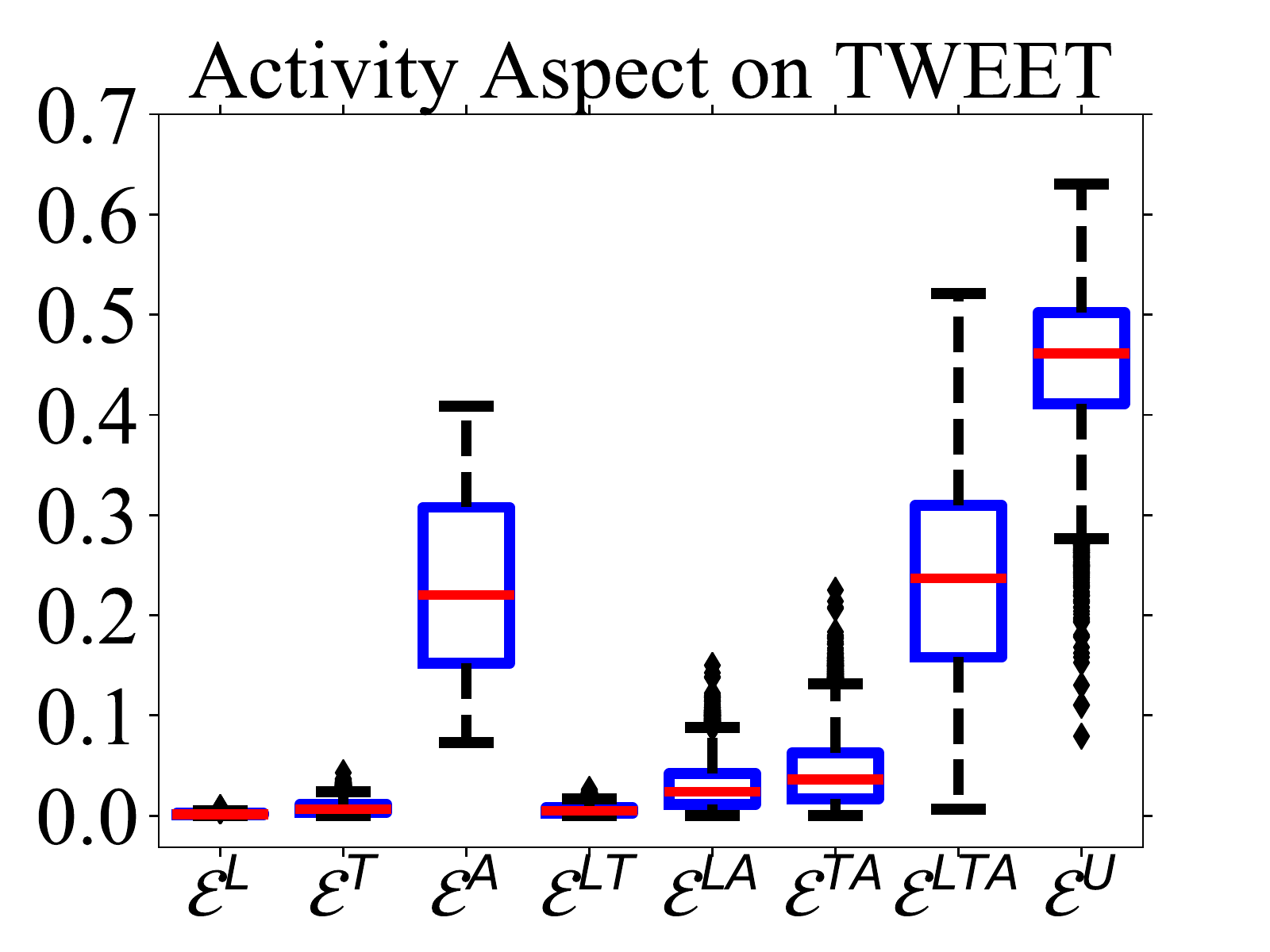}}
	\caption{The attention distributions of each type of hyperedges in three aspects on Telecom and TWEET datasets.} 
	\label{fig::atten}
\end{figure*}

\subsection{Ablation Study on Model Design}
\label{sec:4.3}

In this section, 
we illustrate the effectiveness of key designs in DisenHCN.

\subsubsection{Effectiveness of Multi-type Hyperedges}
We propose eight types of hyperedges in heterogeneous hypergraph,
\textit{i.e.},
$\{\mathcal{E}^{\text{L}},
\mathcal{E}^{\text{T}},
\mathcal{E}^{\text{A}},
\mathcal{E}^{\text{LT}},
\mathcal{E}^{\text{LA}},
\mathcal{E}^{\text{TA}},
\mathcal{E}^{\text{LTA}},
\mathcal{E}^{\text{U}}
\}$,
to capture the high-order relations of multiple user similarities and user  interactions. To evaluate the impact of those types of hyperedges, we compare the performance of the models without a certain type of hyperedges. Specifically, we compare the models removing specific one hyperedge in $\{\mathcal{E}^{\text{L}},
\mathcal{E}^{\text{T}},
\mathcal{E}^{\text{A}},
\mathcal{E}^{\text{LT}},
\mathcal{E}^{\text{LA}},
\mathcal{E}^{\text{TA}},
\mathcal{E}^{\text{LTA}},
\mathcal{E}^{\text{U}}
\}$.

Table~\ref{tab::ablation} shows the results on the four datasets. The results
show that removing any type of hyperedge will lead to performance drop, which verifies the effectiveness of the designed multi-type hyperedges.
Furthermore, DisenHCN suffers a huge performance drop without $\mathcal{E}^{\text{U}}$, which demonstrates the modeling of complex matching between user and spatiotemporal activity via hyperedge is of great importance. 
For the hyperedges that capture the distinct user similarities, we can observe that DisenHCN performs the worst when removing $\mathcal{E}^{\text{LTA}}$. This can be explained that $\mathcal{E}^{\text{LTA}}$ captures the most fine-grained user similarities and contributes most to the final performance.

\begin{figure*}[t]
	\centering
	\subfloat[Telecom]{\includegraphics[width=0.48\columnwidth]{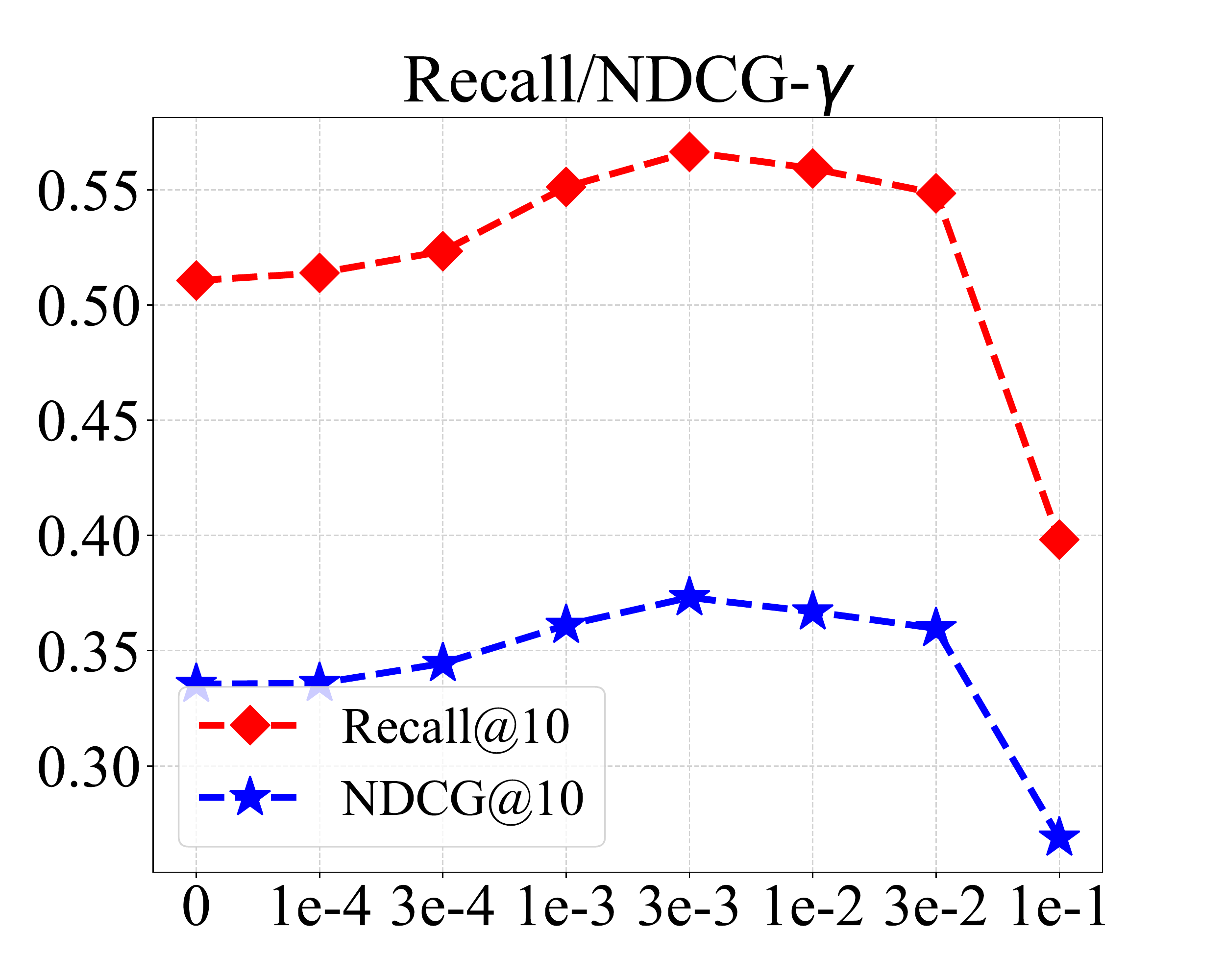}}
	\subfloat[TalkingData]{\includegraphics[width=0.48\columnwidth]{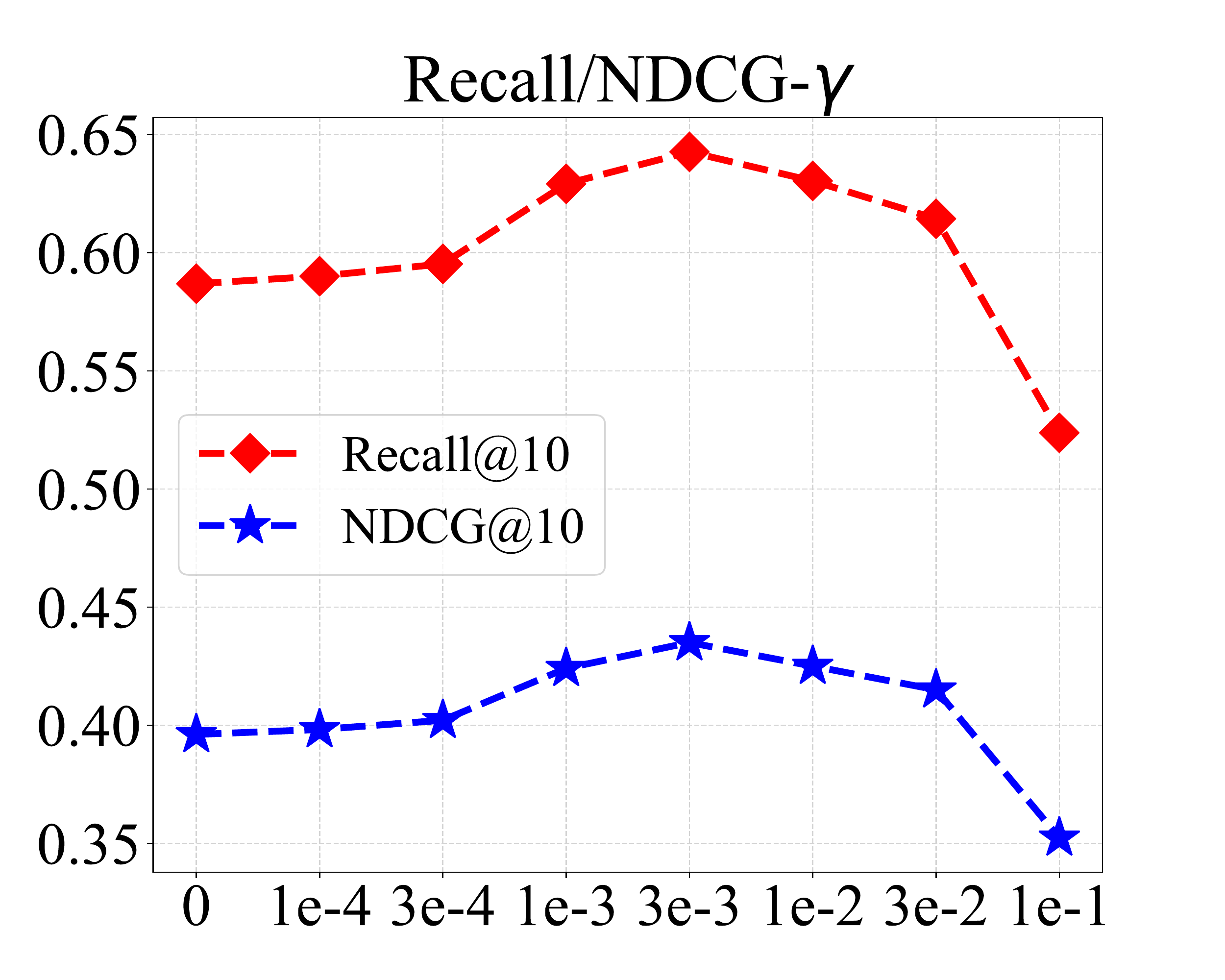}}
	\subfloat[4SQ]{\includegraphics[width=0.48\columnwidth]{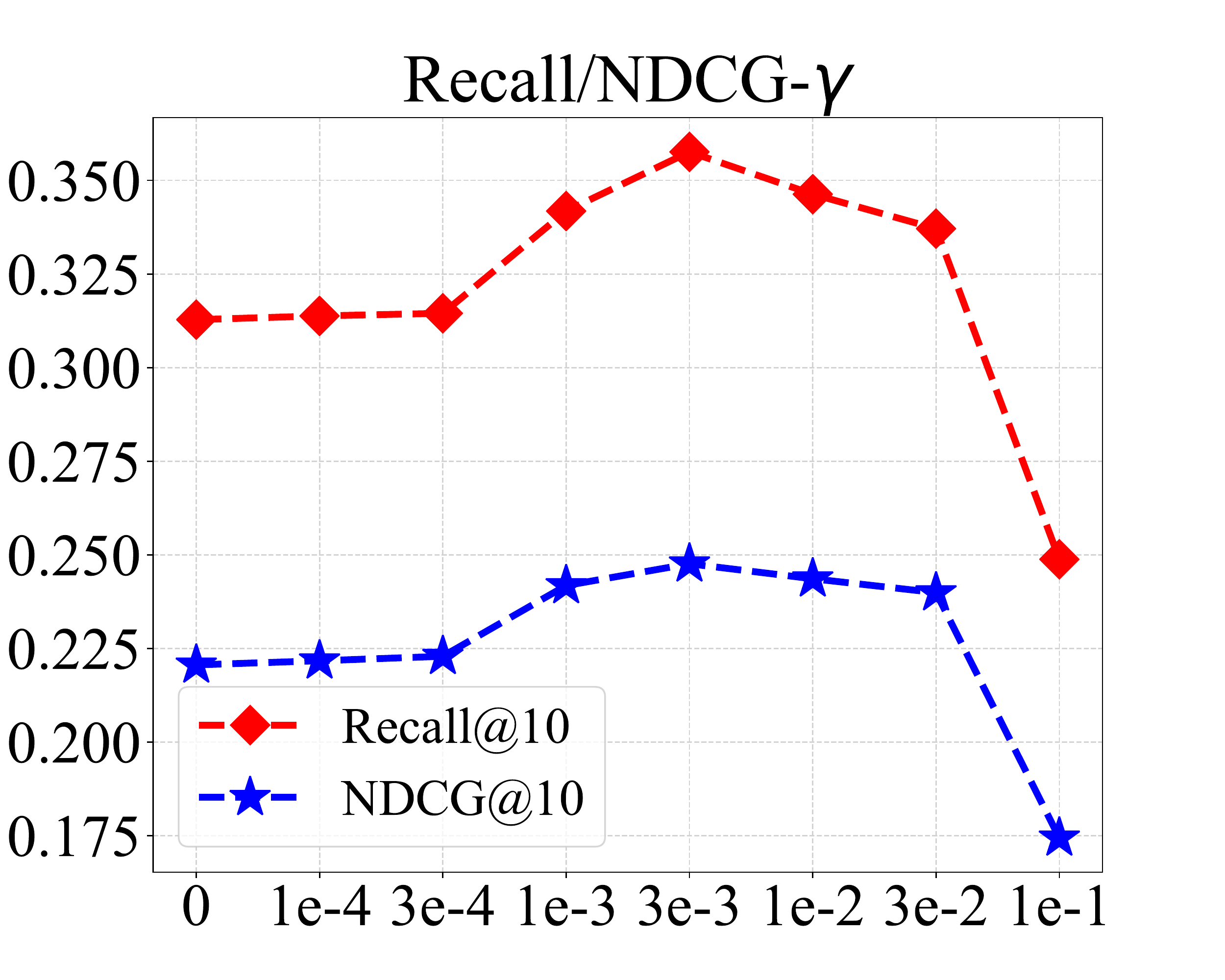}}
	\subfloat[TWEET]{\includegraphics[width=0.48\columnwidth]{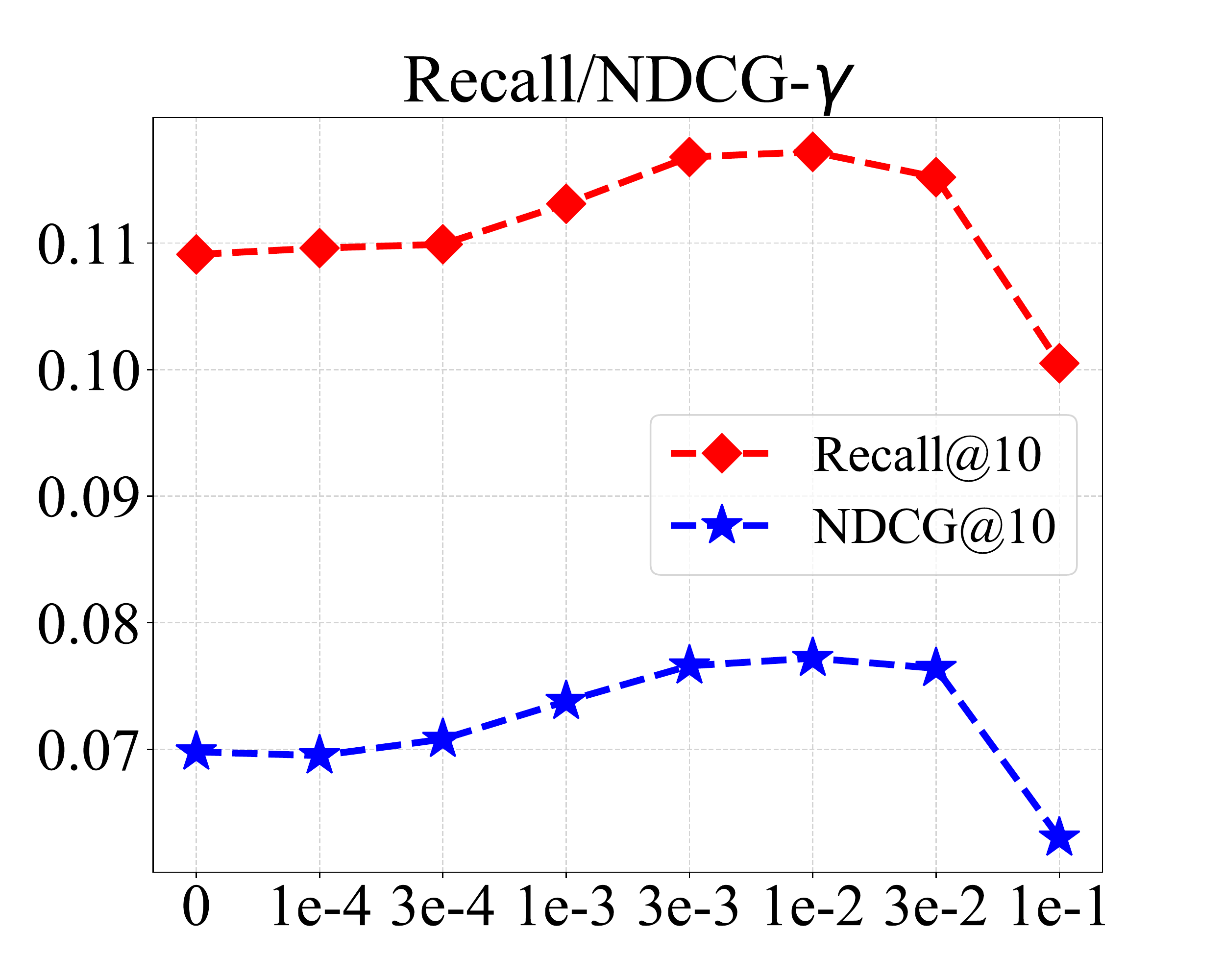}}
	\caption{Impact of independent modeling coefficient $\gamma$ on four datasets.} 
	\label{fig::lambda}
		\vspace{-0.4cm}
\end{figure*}

\begin{figure*}[t]
	\centering
	\subfloat[Telecom]{\includegraphics[width=0.48\columnwidth]{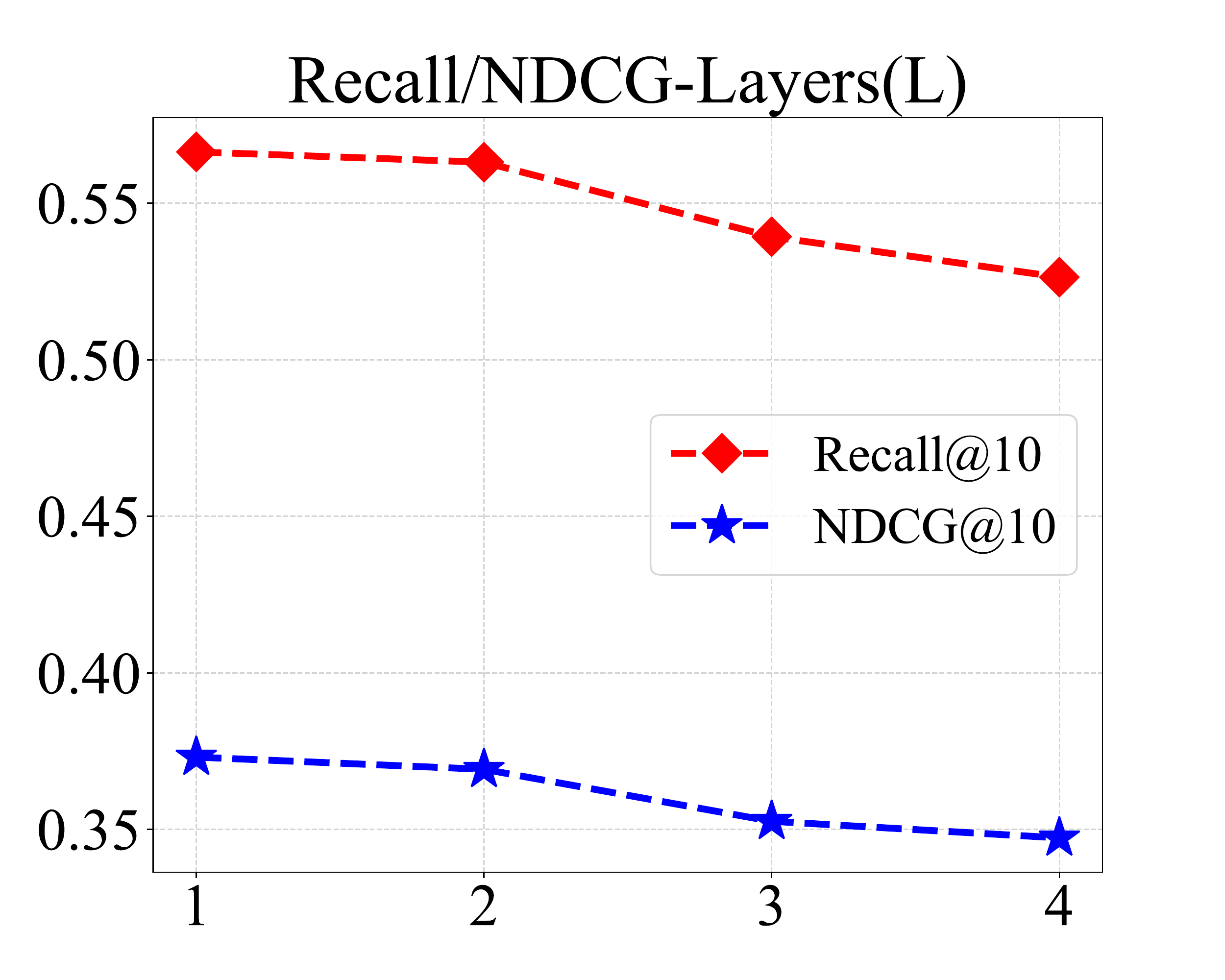}}
	\subfloat[TalkingData]{\includegraphics[width=0.48\columnwidth]{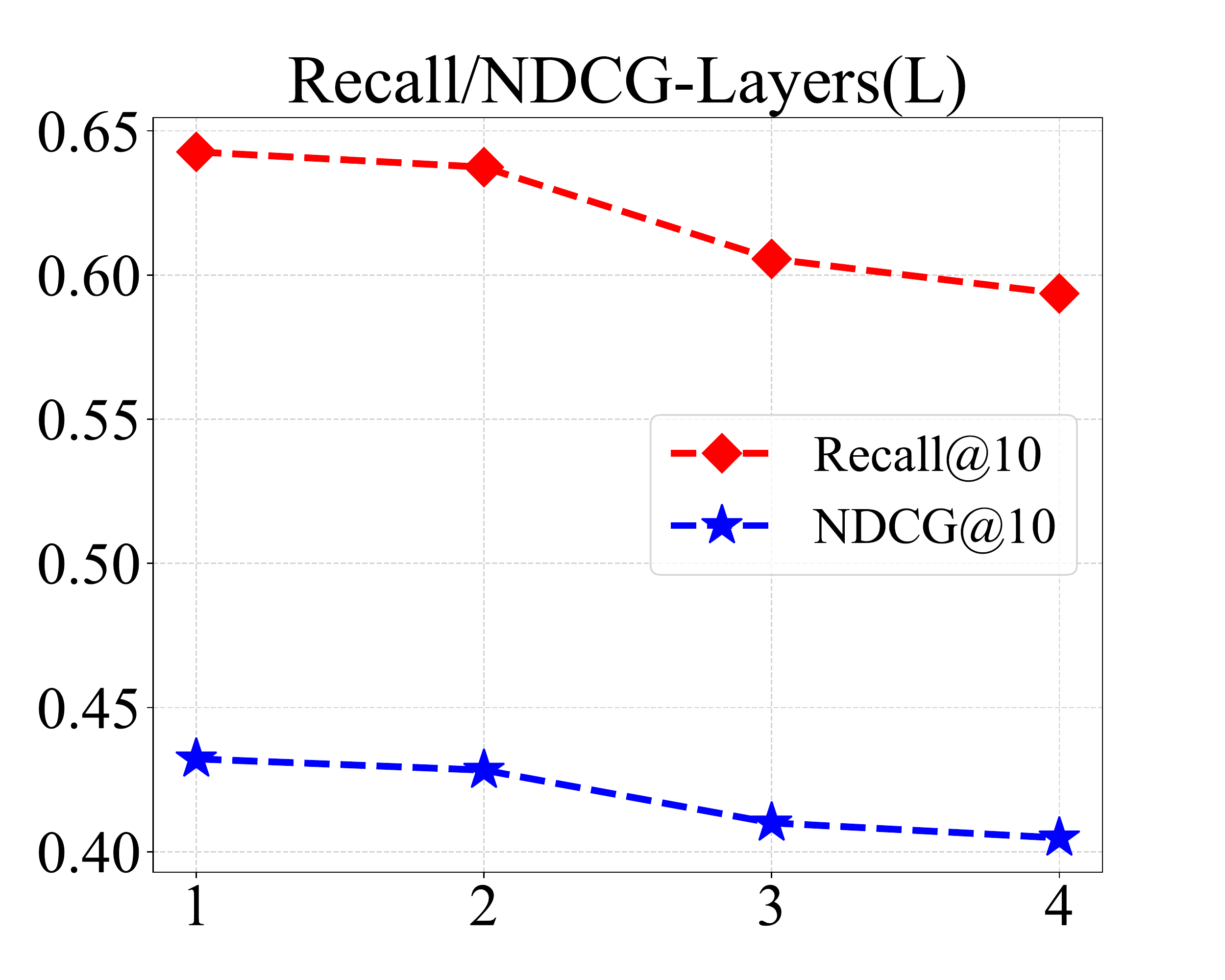}}
	\subfloat[4SQ]{\includegraphics[width=0.48\columnwidth]{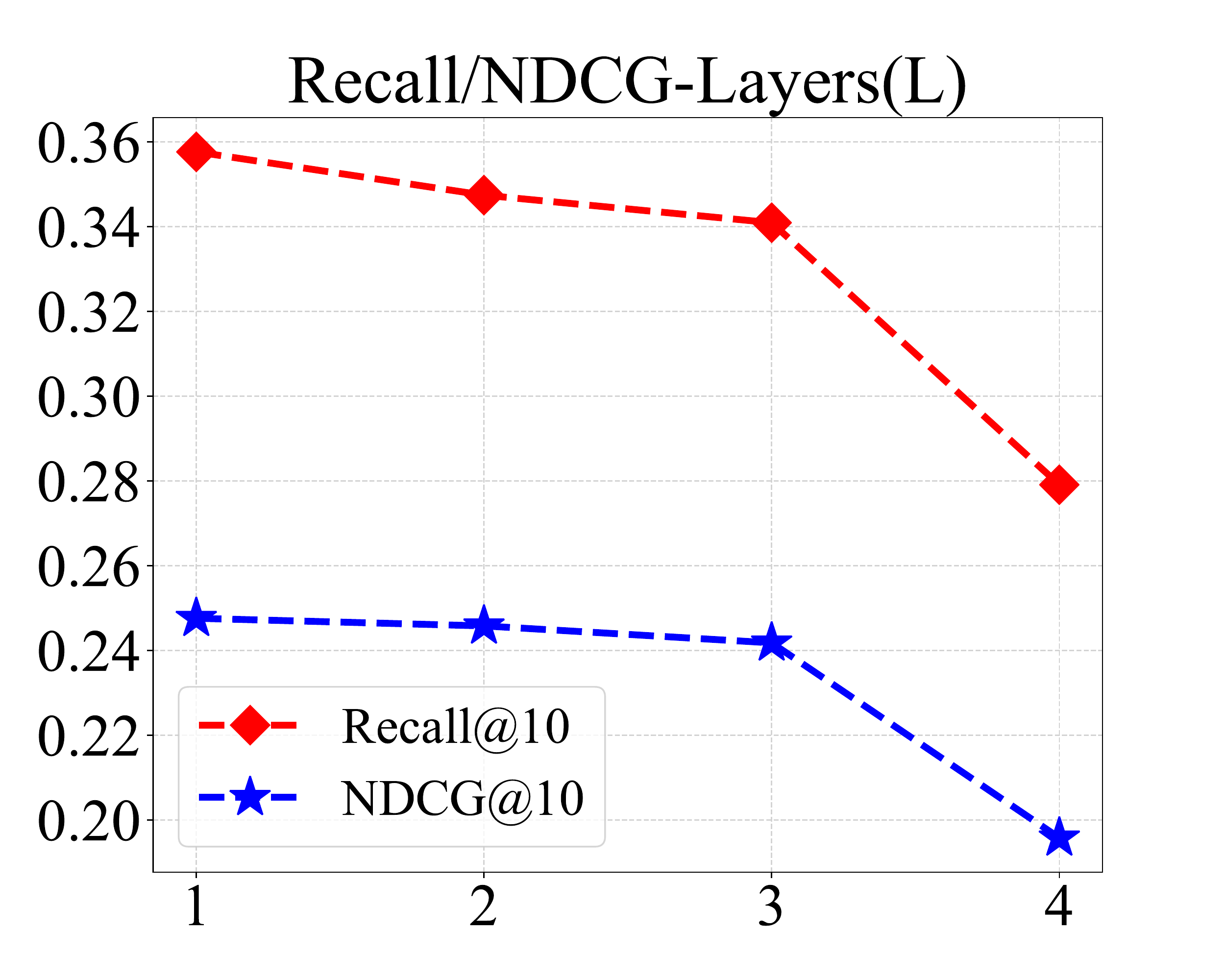}}
	\subfloat[TWEET]{\includegraphics[width=0.48\columnwidth]{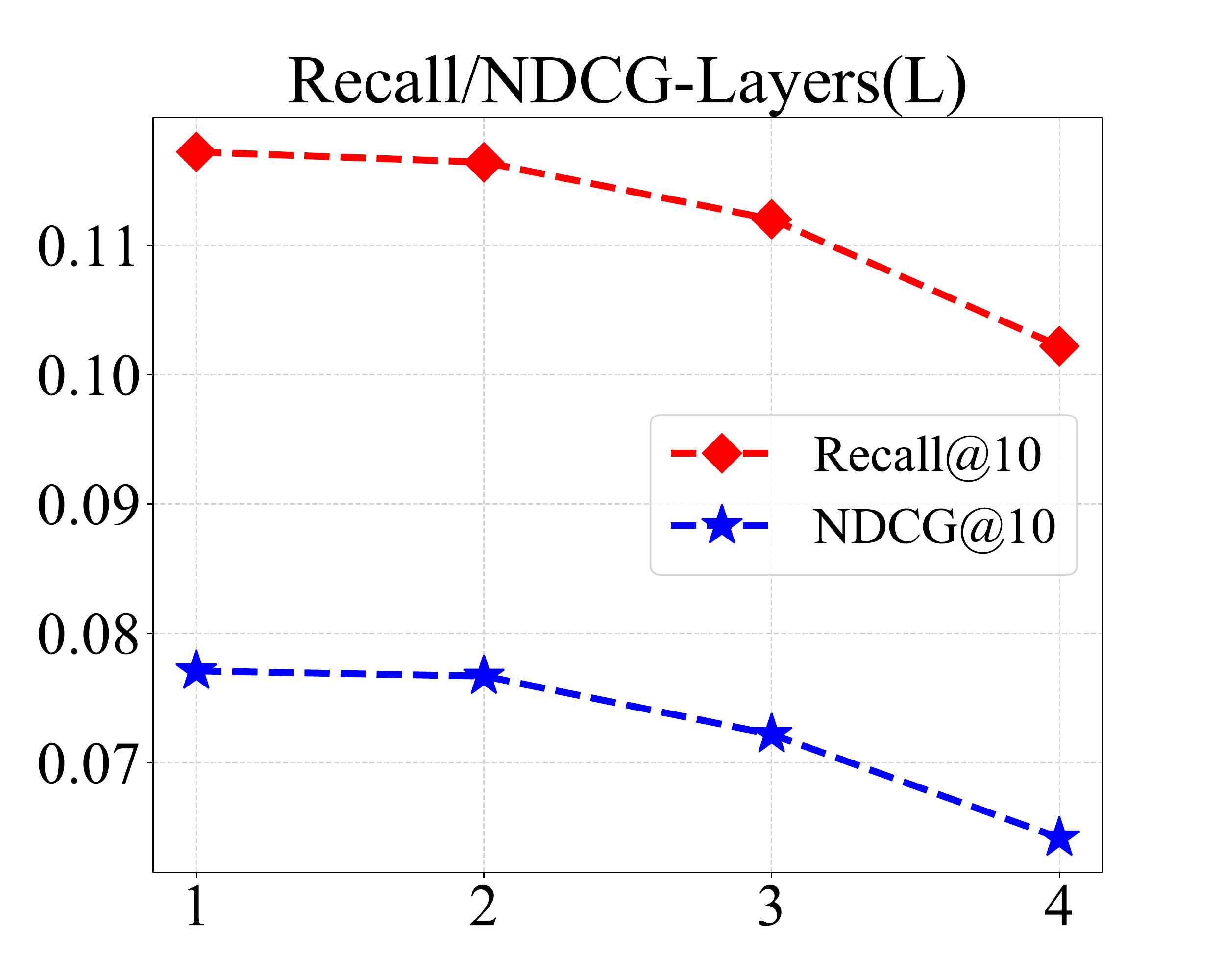}}
	\caption{Impact of the depth of DisenHCN $L$ on four datasets.} 
	\label{fig::layer}
		\vspace{-0.4cm}
\end{figure*}

\subsubsection{Effectiveness of Disentangled Embedding}
We project the user embeddings into three disentangled subspaces to model user preferences on multiple aspects, \textit{i.e.}, location, time, and activity.
To verify its effectiveness, we compare the performance of models with and without disentangled transition. As the results in Table~\ref{tab::ablation} show, the performance of DisenHCN decreases significantly when disentangled transition is removed, which demonstrates the effectiveness of our design.

\subsubsection{Effectiveness of Intra-type Aggregation}
In our DisenHCN, we adopt Eff-HGConv unit in the intra-type aggregation in each type of hyperedges. To evaluate its impact, we compare the performance of models with different hypergraph aggregation schemes, \textit{i.e.}, the model with hypergraph convolution~\cite{feng2019HGNN} (HGConv) and the model with proposed Eff-HGConv.

From the results in Table~\ref{tab::ablation}, we can observe that the model with the proposed Eff-HGConv outperforms HGConv. This can be explained that the calculation of hypergraph convolution in Eff-HGConv involves fewer parameters and is easier to converge. These results convincingly verify the effectiveness of our proposed Eff-HGConv unit.

\subsubsection{Effectiveness of Inter-type Aggregation}
In inter-type aggregation,
aggregation is operated among all hyperedge-types in each subspace; learned type-specific user embeddings are further fused with attention mechanism~\cite{vaswani2017attention}. 
To evaluate the effectiveness of our design, we compare the performance with different fusion functions (\textit{i.e.} Max, Mean, and Attention). According to the results in  Table~\ref{tab::ablation}, we can observe that the model with attention achieves better performance than Max and Mean functions. This can be explained that DisenHCN can adaptively fuse the impact from different types of hyperedges to obtain more comprehensive representations.

\subsubsection{Attention distribution analysis}
To further investigate how DisenHCN adaptively fuses the user preference in each aspect on corresponding types of hyperedges with attention, we conduct the attention distribution analysis with a box plot on Telecom and TWEET. Specifically, we show the attention distributions of each type of hyperedges in three aspects (\textit{i.e.} location, time, and activity aspect). From the results in Figure~\ref{fig::atten},
we have the following observations.
\begin{itemize}[leftmargin=*]
	\item \textbf{$\mathcal{E}^U$ contributes the most to disentangled user preferences in all aspects.} The average attention score of $\mathcal{E}^U$ is the largest on all datasets, which means that $\mathcal{E}^U$ contributes the most to user preferences in all aspects. This is consistent with the result in Table~\ref{tab::ablation} and further verifies that modeling user-centered one-to-many interaction is essential.
	\item \textbf{The disentangled user preference is determined by the hyperedges of the corresponding aspect.} For example, when generating the location-aware aspect of disentangled user embedddings (reflecting user's location preference), the location related hyperedges (\textit{i.e.} $\{\mathcal{E}^{L}$, $\mathcal{E}^{LT}$, $\mathcal{E}^{LA}$, $\mathcal{E}^{LTA}$, and $\mathcal{E}^{U}\}$) contribute more than other hyperedges. 
	This validates the effectiveness of the efficient hypergraph convolution layer.
	Our proposed DisenHCN disentangles user preference in three aspects, adaptively selects the useful information in some aspects, and filters out the noise in other aspects by learnable attention weights, resulting in the best performance.
	
\end{itemize}

In summary, the attention distribution analysis demonstrates the necessity of disentangled modeling of user preferences (\textit{i.e.} location-, time- and activity-aware aspects) and further verifies the effectiveness of efficient hypergraph convolution layer.

\begin{figure*}[t]
	\centering
	\subfloat{\includegraphics[width=0.48\columnwidth]{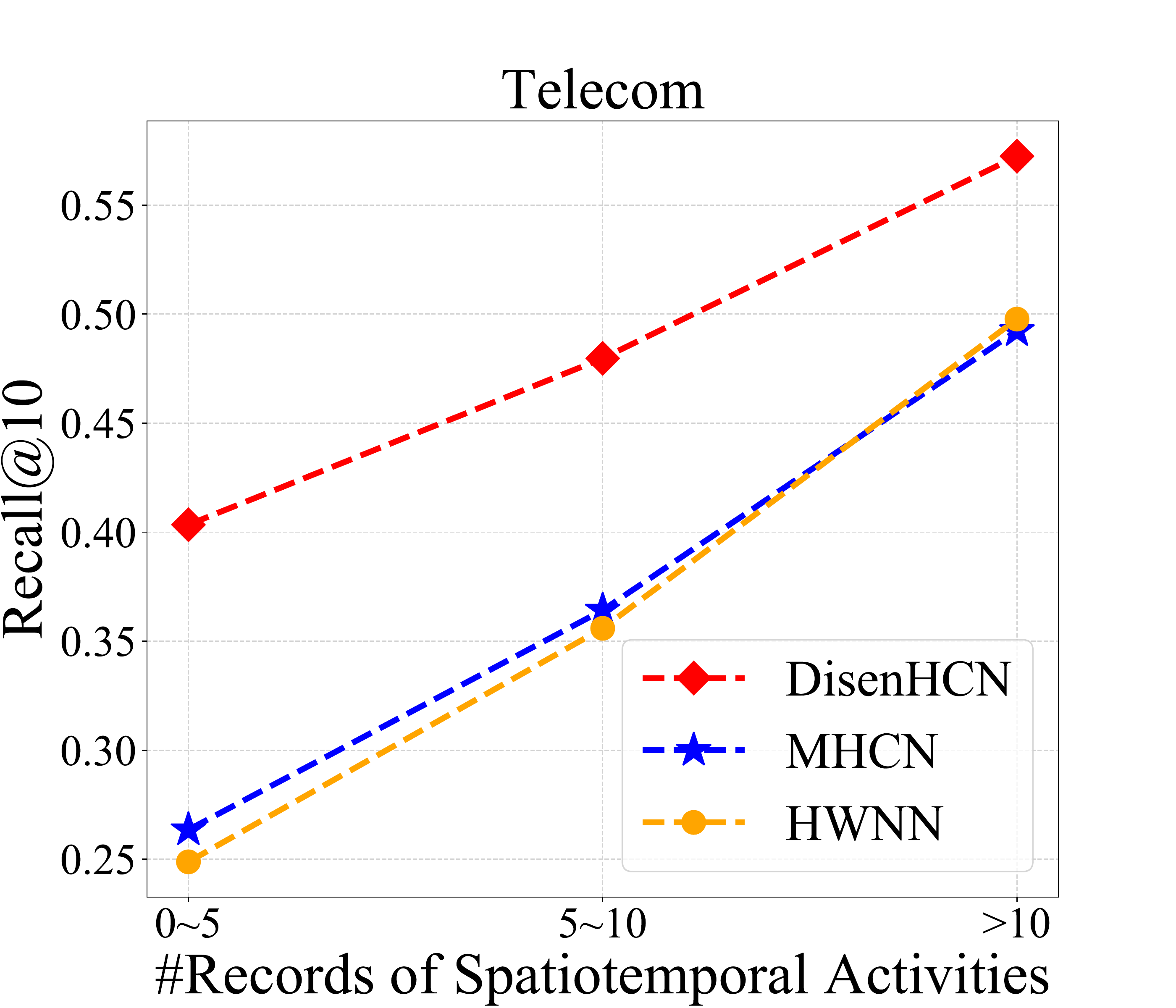}}
	\subfloat{\includegraphics[width=0.48\columnwidth]{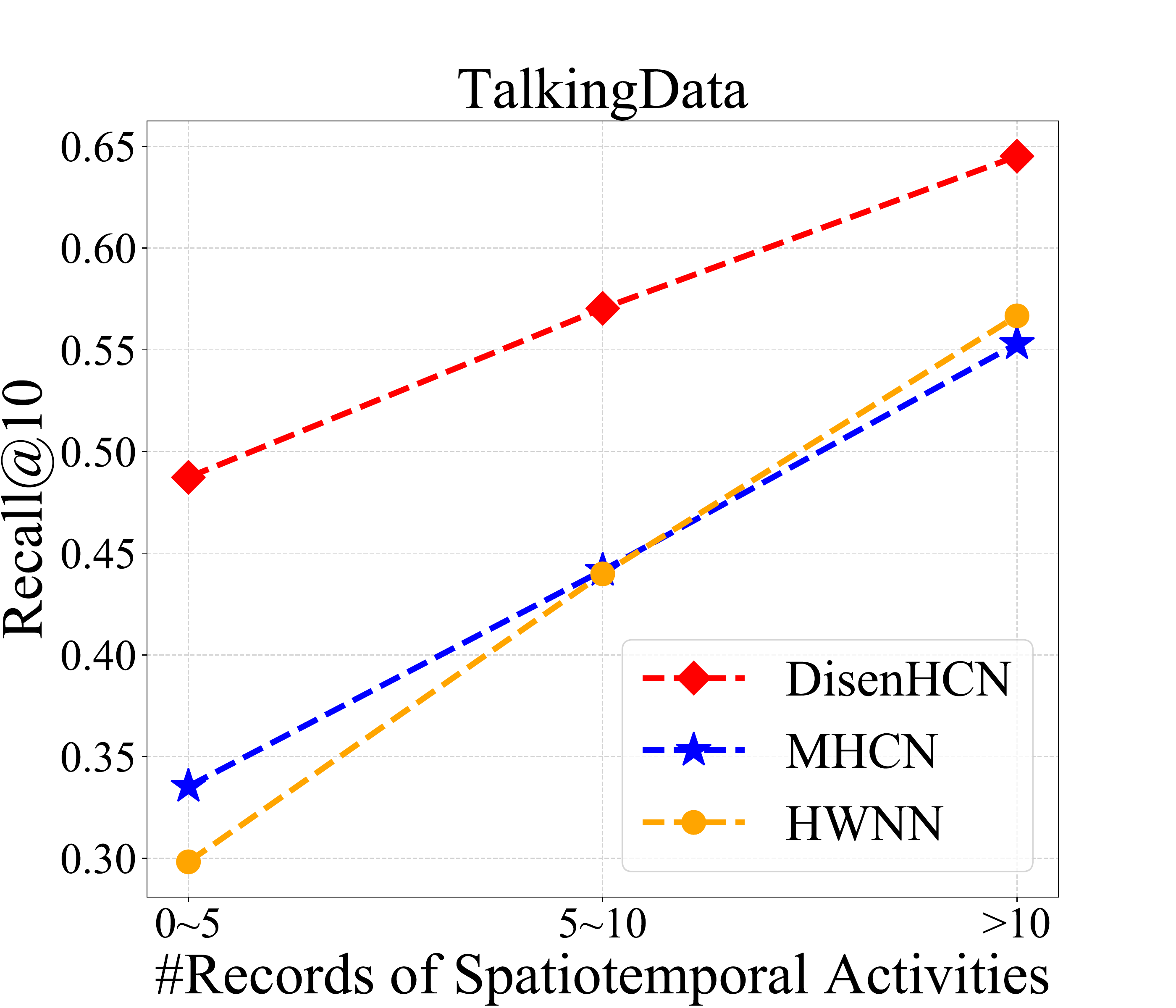}}
	\subfloat{\includegraphics[width=0.48\columnwidth]{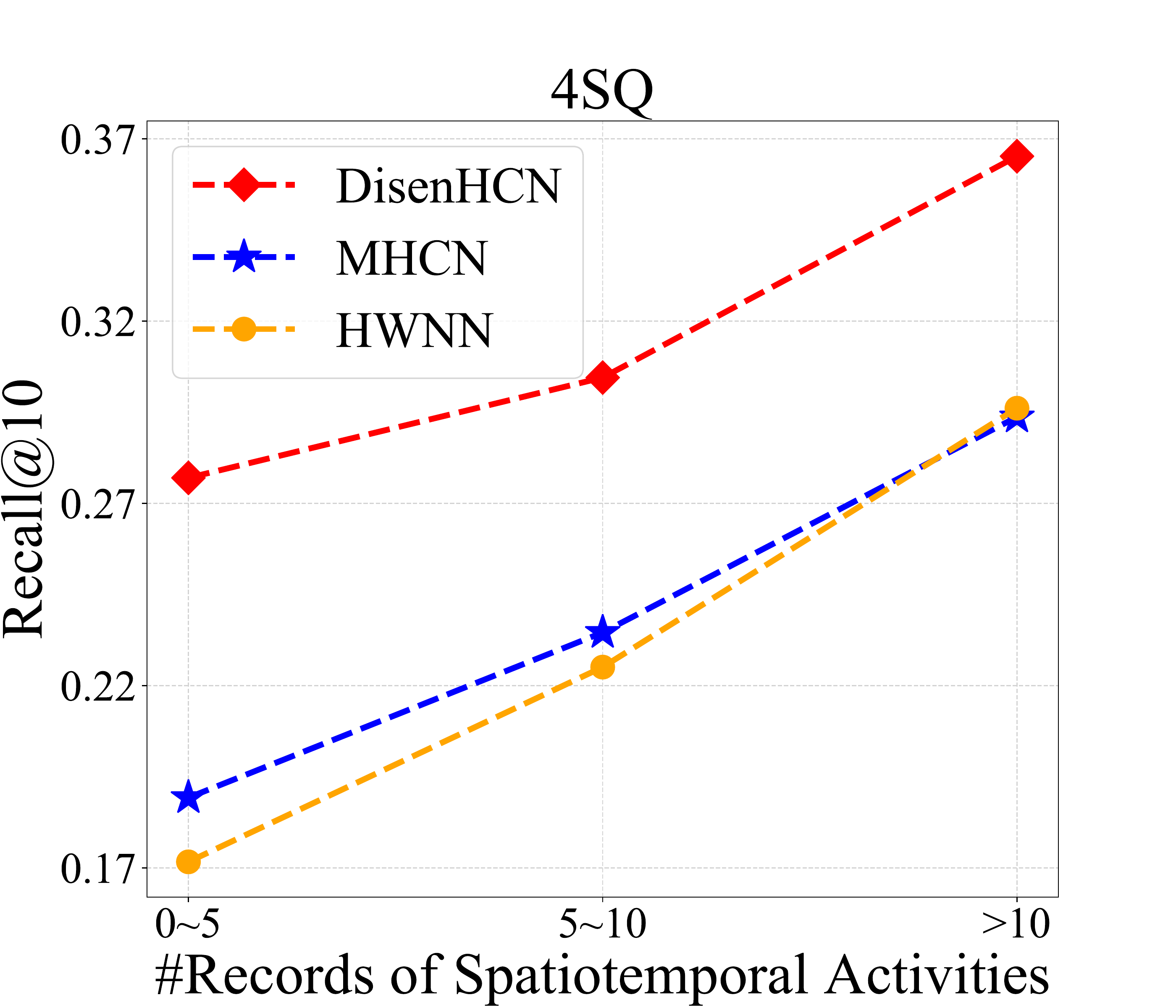}}
	\subfloat{\includegraphics[width=0.48\columnwidth]{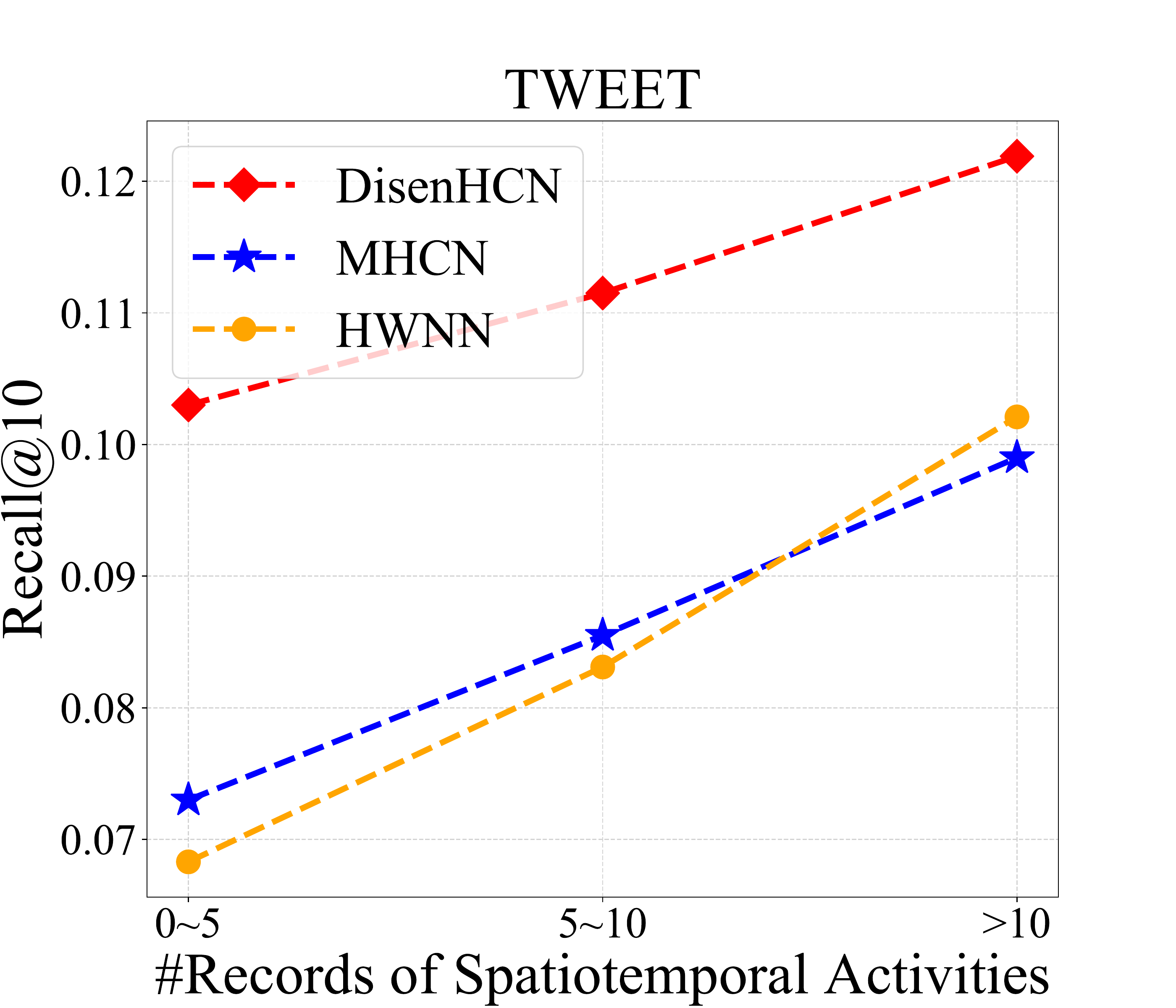}}\\
	\subfloat{\includegraphics[width=0.48\columnwidth]{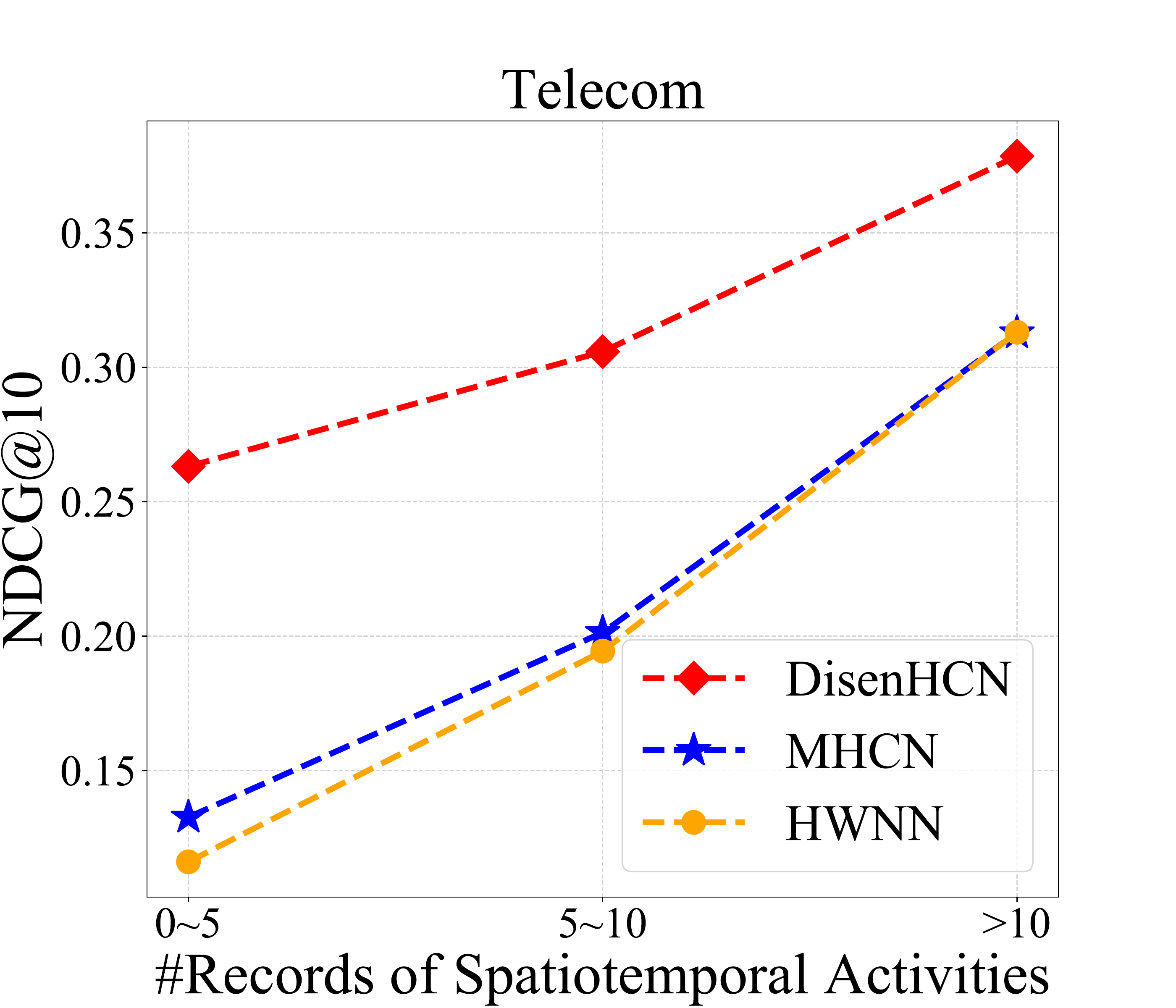}}
	\subfloat{\includegraphics[width=0.48\columnwidth]{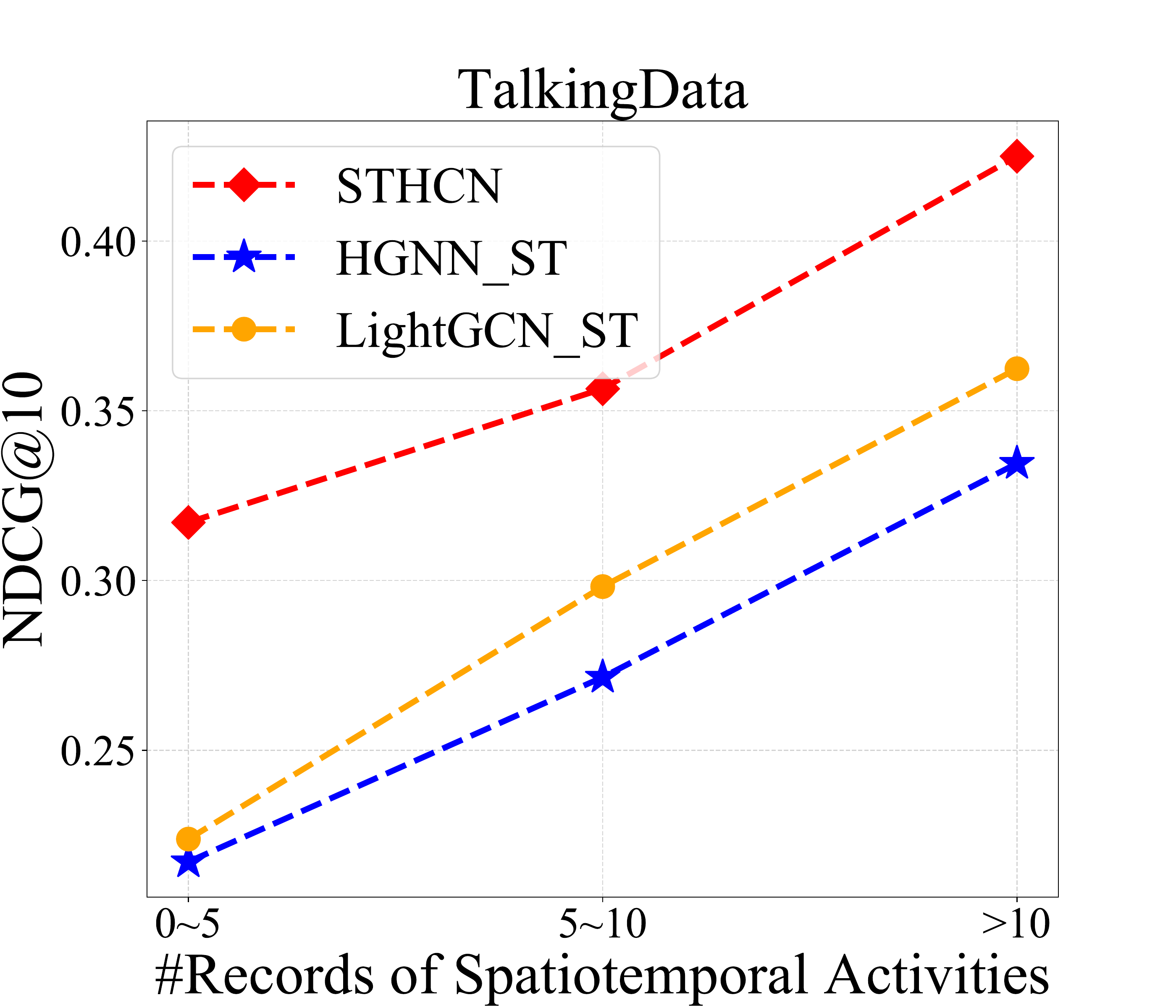}}
	\subfloat{\includegraphics[width=0.48\columnwidth]{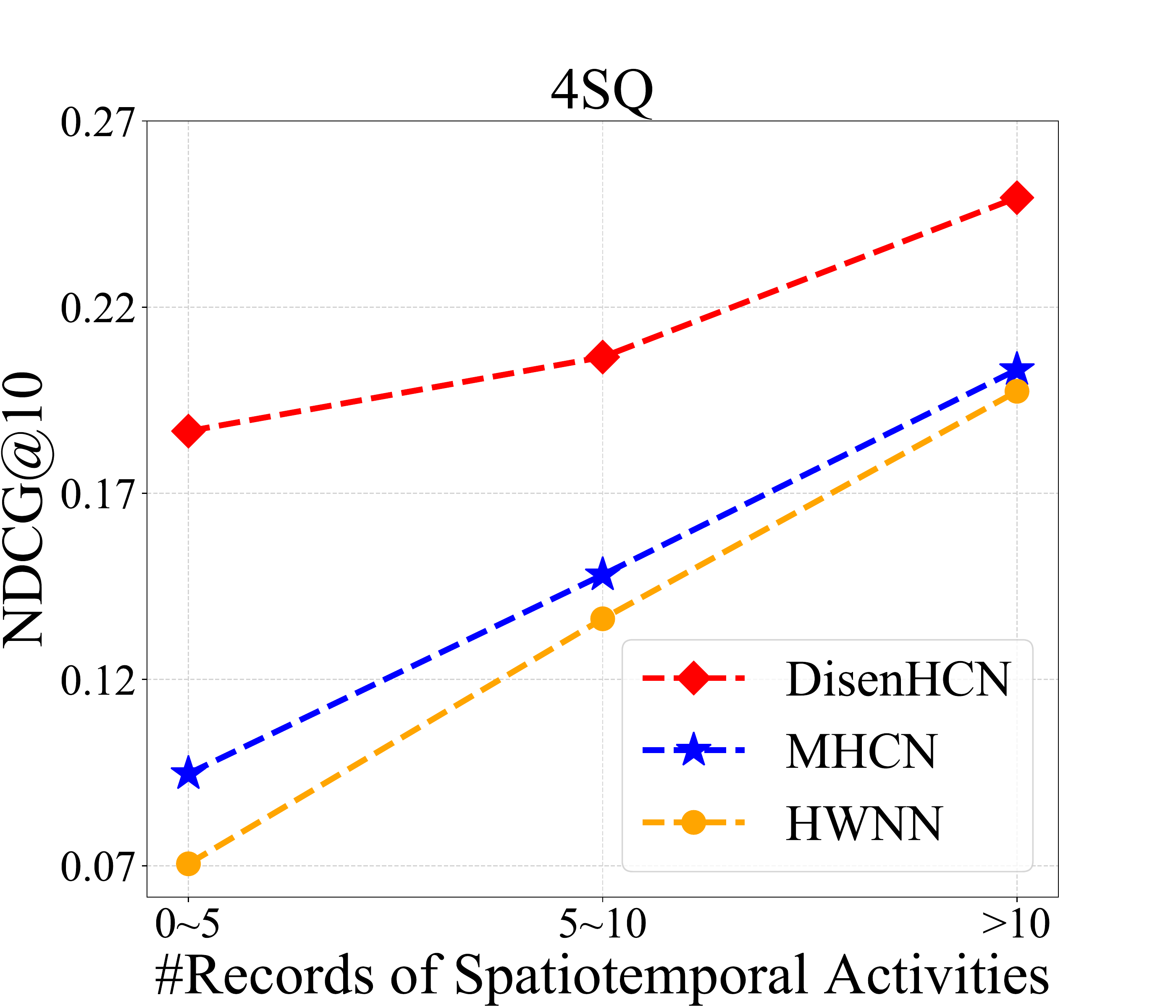}}
	\subfloat{\includegraphics[width=0.48\columnwidth]{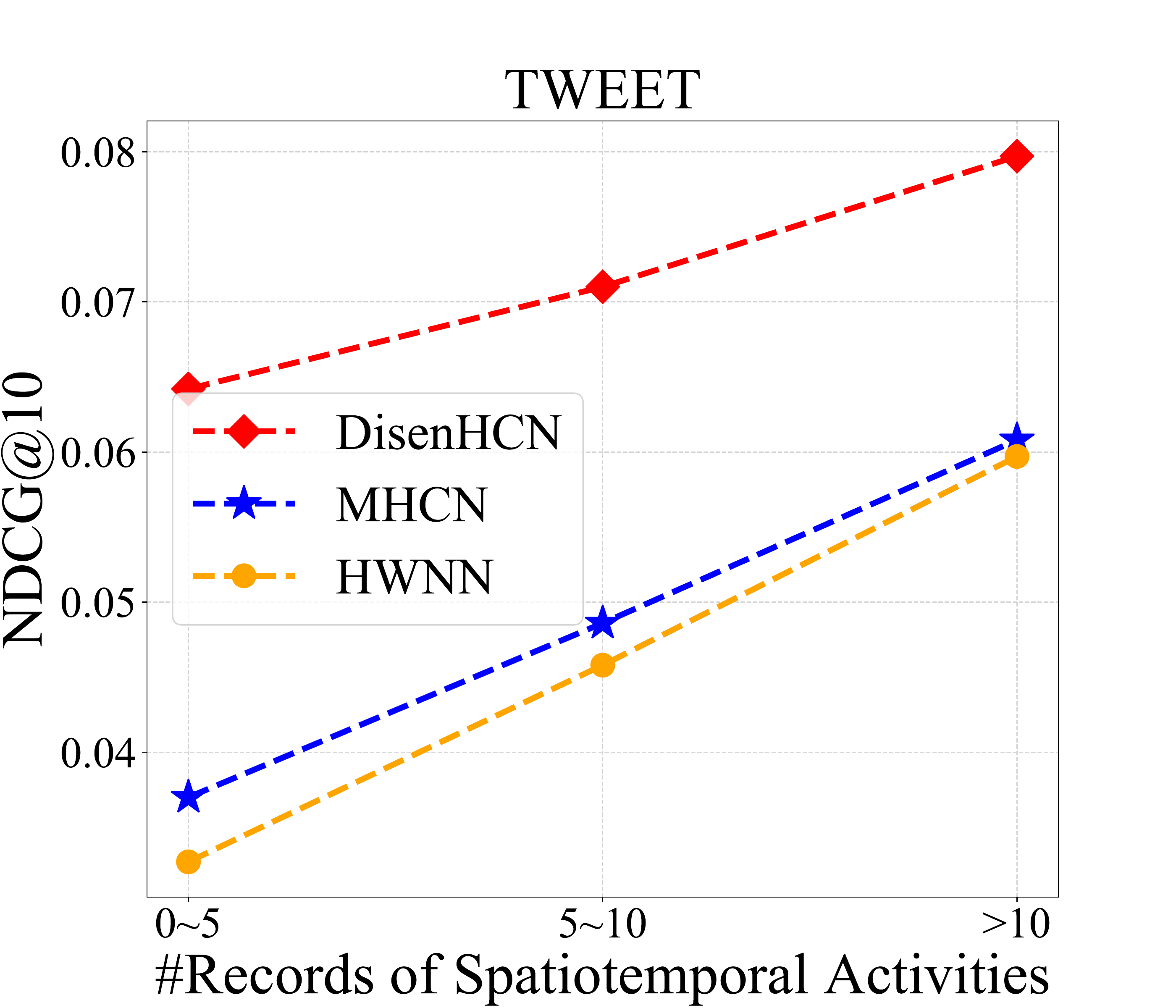}}
	\caption{Performance comparisons with different data sparsity on four datasets.} 
	\label{fig::Telecom_s}
\end{figure*}

\subsection{Sensitivity and Hyper-parameter Study}
\label{sec:hyperpara}
In this section, we investigate the sensitivity of hyper-parameters of our proposed DisenHCN on four datasets.

\subsubsection{Independence-modeling loss coefficient $\gamma$} To further test the effect of coefficient of independence modeling loss, 
$\gamma$ in~\eqref{eqn::total_loss}, we vary it in $\{1e^{-4}, 3e^{-4}, 1e^{-3}, 3e^{-3}, 1e^{-2}, 3e^{-2}, 1e^{-1}\}$. 
According to the results in Fig.~\ref{fig::lambda},
with the increase of $\gamma$, the performances raise first and then remain relatively stable in a certain interval ($1e^{-4} - 3e^{-3}$). The optimal
value for Telecom, TalkingData and 4SQ is $3e^{-3}$ while the optimal
value for TWEET is $1e^{-2}$. The performance will drop significantly when $\gamma$ is larger than $1e^{-2}$.
The coefficient $\gamma$ controls the effect of independence modeling and its value reflects independence-extent of disentangled user embeddings.
When $\gamma$ is relatively small, the disentangled chunks of user embeddings may be similar, which weakens the capability of disentangled chunks. When the values of $\gamma$ are too large, the too strong regularization will make model optimization more difficult and worsen the final performance.
\rev{Note that the minor differences in the same interval (\textit{i.e.}, at the same order of magnitude) have little effect on the performance in our experiments. To reduce the search cost, we choose two promising values, 1 and 3. Results in Fig.~\ref{fig::lambda} also show that the performance of our model remains relatively stable in an interval with the same order of magnitude, such as $[1e^{-3}, 10e^{-3}]$.}

\subsubsection{Number of aggregation layers $L$} 
To investigate how the depth of aggregation layers, 
affects the performance, we vary $L$ in $\{1,2,3,4\}$. As the results in Fig.~\ref{fig::layer} show, our DisenHCN achieves the best performance when the depth is set to 1. When we increase the number of layers, the performance of DisenHCN drops on all datasets, which is also found in the existing hypergraph models~\cite{LCFN,ji2020dual,yu2021MHCN} on recommendation. Notably, most GCN-based models achieve the best performance with two layers, which is a trade-off of performance gain from high-order information (reach k-hop neighbors by stacking k layers) and performance degradation from over-smoothing. However, our DisenHCN has fully captured the high-order relations with hypergraph, which means stacking multiple layers will only lead to over-smoothing rather than additional performance improvement. That is the reason why DisenHCN usually achieves the best performance with only one layer. Undoubtedly, our proposed DisenHCN is more efficient and has the potential to be applied in real-world scenarios with large-scale and sparse data, since there is no necessity of stacking multiple layers for DisenHCN to capture high-order relations.

\subsubsection{Impact of Data Sparsity}\label{sec::sparse}
Data sparsity is a critical issue for spatiotemporal activity prediction, since users' activities can always be few compared with the full space. 
Different from most existing models, our proposed DisenHCN explores multi-type and multi-grained user-similarities via hyperedges. 
Owing to this design, our DisenHCN can capture high-order relations, avoiding the insufficient use of the sparse user activities.
To study whether DisenHCN can alleviate the data sparsity issue, we divide users into three groups according to the number of their spatiotemporal activity records. We present the prediction performance on four datasets in Fig.~\ref{fig::Telecom_s}. According to Fig.~\ref{fig::Telecom_s}, we can observe that the performance gap between DisenHCN and the state-of-the-art methods (MHCN and HWNN) becomes larger with sparser data. Even in the user group with \rev{fewer} than five spatiotemporal activity records, DisenHCN still has an excellent performance on all datasets, which demonstrates the strong power of our DisenHCN model to alleviate the data sparsity issue.


\section{Conclusions and Future Work}\label{sec::conclusion}
In this work, we approach the problem of spatiotemporal activity prediction with hypergraph and propose a disentangled hypergraph convolutional network-based model, DisenHCN. Specifically, we first construct a heterogeneous hypergraph with multiple types of hyperedges to capture the multi-type and multi-grained user similarities and complex matching between user and spatiotemporal activity. To model the users' entangled preferences, we project users to disentangled embeddings for decoupling preferences among location, time and activity. 
Finally, we fuse the information in each aspect with the proposed efficient hypergraph convolution unit (Eff-HGConv) in both intra-type and inter-type manners.
Extensive experiments on four real-world datasets demonstrate the effectiveness and superiority of our DisenHCN. 
For future work, we plan to deploy our method to real-world applications and conduct online A/B tests to further evaluate the performance.

\section*{Acknowledgment}
This work is supported in part by National Natural Science Foundation of China under 61971267, 61972223 and U20B2060.
This work is also supported in part by International Postdoctoral Exchange Fellowship Program (Talent-Introduction Program) under YJ20210274.

\bibliographystyle{IEEEtran}
\balance
\bibliography{bibliography}

\begin{thebibliography}{10}
\providecommand{\url}[1]{#1}
\csname url@samestyle\endcsname
\providecommand{\newblock}{\relax}
\providecommand{\bibinfo}[2]{#2}
\providecommand{\BIBentrySTDinterwordspacing}{\spaceskip=0pt\relax}
\providecommand{\BIBentryALTinterwordstretchfactor}{4}
\providecommand{\BIBentryALTinterwordspacing}{\spaceskip=\fontdimen2\font plus
\BIBentryALTinterwordstretchfactor\fontdimen3\font minus
  \fontdimen4\font\relax}
\providecommand{\BIBforeignlanguage}[2]{{%
\expandafter\ifx\csname l@#1\endcsname\relax
\typeout{** WARNING: IEEEtran.bst: No hyphenation pattern has been}%
\typeout{** loaded for the language `#1'. Using the pattern for}%
\typeout{** the default language instead.}%
\else
\language=\csname l@#1\endcsname
\fi
#2}}
\providecommand{\BIBdecl}{\relax}
\BIBdecl

\bibitem{w4kdd13}
Q.~Yuan, G.~Cong, Z.~Ma, A.~Sun, and N.~M. Thalmann, ``Who, where, when and
  what: discover spatio-temporal topics for twitter users,'' in \emph{KDD},
  2013, pp. 605--613.

\bibitem{WDGTC}
Z.~Li, N.~D. Sergin, H.~Yan, C.~Zhang, and F.~Tsung, ``Tensor completion for
  weakly-dependent data on graph for metro passenger flow prediction,'' in
  \emph{AAAI}, vol.~34, 2020, pp. 4804--4810.

\bibitem{crossmapwww17}
C.~Zhang, K.~Zhang, Q.~Yuan, H.~Peng, Y.~Zheng, T.~Hanratty, S.~Wang, and
  J.~Han, ``Regions, periods, activities: Uncovering urban dynamics via
  cross-modal representation learning,'' in \emph{WWW}, 2017, pp. 361--370.

\bibitem{zhang2016gmove}
C.~Zhang, K.~Zhang, Q.~Yuan, L.~Zhang, T.~Hanratty, and J.~Han, ``Gmove:
  Group-level mobility modeling using geo-tagged social media,'' in \emph{KDD},
  2016, pp. 1305--1314.

\bibitem{yuan2017pred}
Q.~Yuan, W.~Zhang, C.~Zhang, X.~Geng, G.~Cong, and J.~Han, ``Pred: Periodic
  region detection for mobility modeling of social media users,'' in
  \emph{WSDM}, 2017, pp. 263--272.

\bibitem{liu2020ACTOR}
Y.~Liu, X.~Ao, L.~Dong, C.~Zhang, and J.~Wang, ``Spatiotemporal activity
  modeling via hierarchical cross-modal embedding,'' \emph{TKDE}, 2020.

\bibitem{zheng2010UCLAF}
V.~W. Zheng, B.~Cao, Y.~Zheng, X.~Xie, and Q.~Yang, ``Collaborative filtering
  meets mobile recommendation: A user-centered approach.'' in \emph{AAAI},
  vol.~10.\hskip 1em plus 0.5em minus 0.4em\relax Citeseer, 2010, pp. 236--241.

\bibitem{MCTF_WWW2015}
P.~Bhargava, T.~Phan, J.~Zhou, and J.~Lee, ``Who, what, when, and where:
  Multi-dimensional collaborative recommendations using tensor factorization on
  sparse user-generated data,'' in \emph{WWW}, 2015, pp. 130--140.

\bibitem{fan2019personalized}
Y.~Fan, Z.~Tu, Y.~Li, X.~Chen, H.~Gao, L.~Zhang, L.~Su, and D.~Jin,
  ``Personalized context-aware collaborative online activity prediction,''
  \emph{UbiComp / ISWC}, vol.~3, no.~4, pp. 1--28, 2019.

\bibitem{yu2020SAGCN}
Y.~Yu and et~al, ``Semantic-aware spatio-temporal app usage representation via
  graph convolutional network,'' \emph{UbiComp / ISWC}, vol.~4, no.~3, pp.
  1--24, 2020.

\bibitem{zhang2018crossmodal}
C.~Zhang and et~al, ``Spatiotemporal activity modeling under data scarcity: A
  graph-regularized cross-modal embedding approach,'' in \emph{AAAI}, vol.~32,
  2018.

\bibitem{li2020apps}
T.~Li and et~al, ``” what apps did you use?”: Understanding the long-term
  evolution of mobile app usage,'' in \emph{WWW}, 2020, pp. 66--76.

\bibitem{he2017NCF}
X.~He, L.~Liao, H.~Zhang, L.~Nie, X.~Hu, and T.-S. Chua, ``Neural collaborative
  filtering,'' in \emph{WWW}, 2017, pp. 173--182.

\bibitem{wang2020DGCF}
X.~Wang, H.~Jin, A.~Zhang, X.~He, T.~Xu, and T.-S. Chua, ``Disentangled graph
  collaborative filtering,'' in \emph{Proceedings of the 43rd International ACM
  SIGIR Conference on Research and Development in Information Retrieval}, 2020,
  pp. 1001--1010.

\bibitem{he2020lightgcn}
X.~He, K.~Deng, X.~Wang, Y.~Li, Y.~Zhang, and M.~Wang, ``Lightgcn: Simplifying
  and powering graph convolution network for recommendation,'' \emph{arXiv
  preprint arXiv:2002.02126}, 2020.

\bibitem{sizov2010geofolk}
S.~Sizov, ``Geofolk: latent spatial semantics in web 2.0 social media,'' in
  \emph{WSDM}, 2010, pp. 281--290.

\bibitem{shashua2005tensorfact}
A.~Shashua and T.~Hazan, ``Non-negative tensor factorization with applications
  to statistics and computer vision,'' in \emph{Proceedings of the 22nd
  international conference on Machine learning}, 2005, pp. 792--799.

\bibitem{chen2019cap}
X.~Chen and et~al, ``Cap: Context-aware app usage prediction with heterogeneous
  graph embedding,'' \emph{UbiComp / ISWC}, vol.~3, no.~1, pp. 1--25, 2019.

\bibitem{GCN}
T.~N. Kipf and M.~Welling, ``Semi-supervised classification with graph
  convolutional networks,'' \emph{arXiv preprint arXiv:1609.02907}, 2016.

\bibitem{icml2006hypergraph}
S.~Agarwal, K.~Branson, and S.~Belongie, ``Higher order learning with graphs,''
  in \emph{ICML}, 2006, pp. 17--24.

\bibitem{yu2021MHCN}
J.~Yu, H.~Yin, J.~Li, Q.~Wang, N.~Q.~V. Hung, and X.~Zhang, ``Self-supervised
  multi-channel hypergraph convolutional network for social recommendation,''
  \emph{arXiv e-prints}, pp. arXiv--2101, 2021.

\bibitem{ji2020dual}
S.~Ji, Y.~Feng, R.~Ji, X.~Zhao, W.~Tang, and Y.~Gao, ``Dual channel hypergraph
  collaborative filtering,'' in \emph{KDD}, 2020, pp. 2020--2029.

\bibitem{ding2020HyperGAT}
K.~Ding, J.~Wang, J.~Li, D.~Li, and H.~Liu, ``Be more with less: Hypergraph
  attention networks for inductive text classification,'' \emph{arXiv preprint
  arXiv:2011.00387}, 2020.

\bibitem{feng2019HGNN}
Y.~Feng, H.~You, Z.~Zhang, R.~Ji, and Y.~Gao, ``Hypergraph neural networks,''
  in \emph{AAAI}, vol.~33, 2019, pp. 3558--3565.

\bibitem{yi2020HRCN}
J.~Yi and J.~Park, ``Hypergraph convolutional recurrent neural network,'' in
  \emph{KDD}, 2020, pp. 3366--3376.

\bibitem{Pagerank}
Y.~Takai, A.~Miyauchi, M.~Ikeda, and Y.~Yoshida, ``Hypergraph clustering based
  on pagerank,'' in \emph{KDD}, 2020, pp. 1970--1978.

\bibitem{wang2020HyperRec}
J.~Wang, K.~Ding, L.~Hong, H.~Liu, and J.~Caverlee, ``Next-item recommendation
  with sequential hypergraphs,'' in \emph{SIGIR}, 2020, pp. 1101--1110.

\bibitem{LCFN}
W.~Yu and Z.~Qin, ``Graph convolutional network for recommendation with
  low-pass collaborative filters,'' in \emph{ICML}.\hskip 1em plus 0.5em minus
  0.4em\relax PMLR, 2020, pp. 10\,936--10\,945.

\bibitem{sun2021HWNN}
X.~Sun, H.~Yin, B.~Liu, H.~Chen, J.~Cao, Y.~Shao, and N.~Q. Viet~Hung,
  ``Heterogeneous hypergraph embedding for graph classification,'' in
  \emph{WSDM}, 2021, pp. 725--733.

\bibitem{zhu2016hete_doc}
Y.~Zhu, Z.~Guan, S.~Tan, H.~Liu, D.~Cai, and X.~He, ``Heterogeneous hypergraph
  embedding for document recommendation,'' \emph{Neurocomputing}, vol. 216, pp.
  150--162, 2016.

\bibitem{yang2019hete-social}
D.~Yang, B.~Qu, J.~Yang, and P.~Cudre-Mauroux, ``Revisiting user mobility and
  social relationships in lbsns: a hypergraph embedding approach,'' in
  \emph{The world wide web conference}, 2019, pp. 2147--2157.

\bibitem{bengio2013representation}
Y.~Bengio, A.~Courville, and P.~Vincent, ``Representation learning: A review
  and new perspectives,'' \emph{IEEE transactions on pattern analysis and
  machine intelligence}, vol.~35, no.~8, pp. 1798--1828, 2013.

\bibitem{burgess2018beta-VAE}
C.~P. Burgess, I.~Higgins, A.~Pal, L.~Matthey, N.~Watters, G.~Desjardins, and
  A.~Lerchner, ``Understanding disentangling in \textit{beta}-vae,''
  \emph{arXiv preprint arXiv:1804.03599}, 2018.

\bibitem{chen2018VAE-ISO}
R.~T. Chen, X.~Li, R.~Grosse, and D.~Duvenaud, ``Isolating sources of
  disentanglement in variational autoencoders,'' \emph{arXiv preprint
  arXiv:1802.04942}, 2018.

\bibitem{higgins2016beta}
I.~Higgins, L.~Matthey, A.~Pal, C.~Burgess, X.~Glorot, M.~Botvinick,
  S.~Mohamed, and A.~Lerchner, ``\textit{beta}-vae: Learning basic visual
  concepts with a constrained variational framework,'' 2016.

\bibitem{kim2018disentangling}
H.~Kim and A.~Mnih, ``Disentangling by factorising,'' in \emph{International
  Conference on Machine Learning}.\hskip 1em plus 0.5em minus 0.4em\relax PMLR,
  2018, pp. 2649--2658.

\bibitem{kingma2013VAE}
D.~P. Kingma and M.~Welling, ``Auto-encoding variational bayes,'' \emph{arXiv
  preprint arXiv:1312.6114}, 2013.

\bibitem{ma2019disenrec-nips}
J.~Ma, C.~Zhou, P.~Cui, H.~Yang, and W.~Zhu, ``Learning disentangled
  representations for recommendation,'' \emph{arXiv preprint arXiv:1910.14238},
  2019.

\bibitem{wang2020disenhan}
Y.~Wang, S.~Tang, Y.~Lei, W.~Song, S.~Wang, and M.~Zhang, ``Disenhan:
  Disentangled heterogeneous graph attention network for recommendation,'' in
  \emph{Proceedings of the 29th ACM International Conference on Information \&
  Knowledge Management}, 2020, pp. 1605--1614.

\bibitem{zheng2021DICE}
Y.~Zheng, C.~Gao, X.~Li, X.~He, Y.~Li, and D.~Jin, ``Disentangling user
  interest and conformity for recommendation with causal embedding,'' in
  \emph{Proceedings of the Web Conference 2021}, 2021, pp. 2980--2991.

\bibitem{scarselli2008graph}
F.~Scarselli, M.~Gori, A.~C. Tsoi, M.~Hagenbuchner, and G.~Monfardini, ``The
  graph neural network model,'' \emph{IEEE transactions on neural networks},
  vol.~20, no.~1, pp. 61--80, 2008.

\bibitem{hamilton2017inductive}
W.~L. Hamilton, R.~Ying, and J.~Leskovec, ``Inductive representation learning
  on large graphs,'' in \emph{NIPS}, 2017, pp. 1025--1035.

\bibitem{wu2020comprehensive}
Z.~Wu, S.~Pan, F.~Chen, G.~Long, C.~Zhang, and S.~Y. Philip, ``A comprehensive
  survey on graph neural networks,'' \emph{IEEE transactions on neural networks
  and learning systems}, vol.~32, no.~1, pp. 4--24, 2020.

\bibitem{gilmer2017neural}
J.~Gilmer, S.~S. Schoenholz, P.~F. Riley, O.~Vinyals, and G.~E. Dahl, ``Neural
  message passing for quantum chemistry,'' in \emph{International conference on
  machine learning}.\hskip 1em plus 0.5em minus 0.4em\relax PMLR, 2017, pp.
  1263--1272.

\bibitem{vaswani2017attention}
A.~Vaswani, N.~Shazeer, N.~Parmar, J.~Uszkoreit, L.~Jones, A.~N. Gomez,
  {\L}.~Kaiser, and I.~Polosukhin, ``Attention is all you need,'' in
  \emph{NeurIPS}, 2017, pp. 5998--6008.

\bibitem{szekely2009brownian}
G.~J. Sz{\'e}kely, M.~L. Rizzo \emph{et~al.}, ``Brownian distance covariance,''
  \emph{The annals of applied statistics}, vol.~3, no.~4, pp. 1236--1265, 2009.

\bibitem{szekely2007measuring}
G.~J. Sz{\'e}kely, M.~L. Rizzo, N.~K. Bakirov \emph{et~al.}, ``Measuring and
  testing dependence by correlation of distances,'' \emph{The annals of
  statistics}, vol.~35, no.~6, pp. 2769--2794, 2007.

\bibitem{2012bpr}
S.~Rendle, C.~Freudenthaler, and et~al., ``Bpr: Bayesian personalized ranking
  from implicit feedback,'' \emph{arXiv preprint arXiv:1205.2618}, 2012.

\bibitem{lowrankMF}
D.~D. Lee and H.~S. Seung, ``Learning the parts of objects by non-negative
  matrix factorization,'' \emph{Nature}, vol. 401, no. 6755, pp. 788--791,
  1999.

\bibitem{Telecom}
D.~Yu and et~al., ``Smartphone app usage prediction using points of interest,''
  \emph{UbiComp / ISWC}, vol.~1, no.~4, pp. 1--21, 2018.

\bibitem{wang2019han}
X.~Wang, H.~Ji, C.~Shi, B.~Wang, Y.~Ye, P.~Cui, and P.~S. Yu, ``Heterogeneous
  graph attention network,'' in \emph{The World Wide Web Conference}, 2019, pp.
  2022--2032.

\bibitem{shi2020odflow}
H.~Shi, Q.~Yao, Q.~Guo, Y.~Li, L.~Zhang, J.~Ye, Y.~Li, and Y.~Liu, ``Predicting
  origin-destination flow via multi-perspective graph convolutional network,''
  in \emph{2020 IEEE 36th International Conference on Data Engineering
  (ICDE)}.\hskip 1em plus 0.5em minus 0.4em\relax IEEE, 2020, pp. 1818--1821.

\bibitem{sawhney2020STHGCN}
R.~Sawhney, S.~Agarwal, A.~Wadhwa, and R.~R. Shah, ``Spatiotemporal hypergraph
  convolution network for stock movement forecasting,'' in \emph{2020 IEEE
  International Conference on Data Mining (ICDM)}.\hskip 1em plus 0.5em minus
  0.4em\relax IEEE, 2020, pp. 482--491.

\bibitem{wang2021apan}
X.~Wang, D.~Lyu, M.~Li, Y.~Xia, Q.~Yang, X.~Wang, X.~Wang, P.~Cui, Y.~Yang,
  B.~Sun \emph{et~al.}, ``Apan: Asynchronous propagation attention network for
  real-time temporal graph embedding,'' in \emph{Proceedings of the 2021
  International Conference on Management of Data}, 2021, pp. 2628--2638.

\bibitem{guo2020dynamic}
K.~Guo, Y.~Hu, Z.~Qian, Y.~Sun, J.~Gao, and B.~Yin, ``Dynamic graph convolution
  network for traffic forecasting based on latent network of laplace matrix
  estimation,'' \emph{IEEE Transactions on Intelligent Transportation Systems},
  2020.

\bibitem{kingma2014adam}
D.~P. Kingma and J.~Ba, ``Adam: A method for stochastic optimization,''
  \emph{arXiv preprint arXiv:1412.6980}, 2014.

\bibitem{glorot2010Xavier}
X.~Glorot and Y.~Bengio, ``Understanding the difficulty of training deep
  feedforward neural networks,'' in \emph{AISTATS}, 2010, pp. 249--256.

\end{thebibliography}

\end{document}